%% file: SRS.tex
\documentclass[lettersize,journal]{IEEEtran}
\usepackage{amsmath,amsfonts}
\usepackage{array}
\usepackage[caption=true,font=footnotesize,labelfont=sf,textfont=sf]{subfig}
\usepackage{textcomp}
\usepackage{stfloats}
\usepackage{url}
\usepackage{tabularx}
\usepackage{verbatim}
\usepackage{graphicx}
\usepackage{cite}
\usepackage{multirow}
\usepackage{color}
\usepackage{caption}
\usepackage{pifont}
\usepackage{booktabs}
\usepackage{multirow}
\usepackage{algorithm, algorithmicx, algpseudocode}
\usepackage{graphicx}
\usepackage{url}
\usepackage{hyperref}
\usepackage[dvipsnames, table]{xcolor}
\usepackage{color}
\definecolor{myellow1}{rgb}{0.94, 0.8, 0.0}
\colorlet{mygray}{white!15}
\colorlet{myellow}{lightgray!15}
\definecolor{l}{RGB}{189, 230, 205}
\definecolor{ll}{RGB}{228,238,188}

\newcommand{\gray}[1]{\textcolor{gray!60}{#1}}

\begin{document}

\title{SRS: Siamese Reconstruction-Segmentation Network based on Dynamic-Parameter Convolution}
% \title{DPConv: Dynamic-Parameter Convolution with Pre-training Reconstruction for segmentation task}

\author{Bingkun Nian{\dag}, Fenghe Tang{\dag}, Jianrui Ding, Jie Yang, Zhonglong Zheng, Shaohua Kevin Zhou{$^{*}$}, Wei Liu{$^{*}$}

\thanks{\dag \   These authors contributed equally. $^{*}$ Corresponding author: S. Kevin Zhou  and Wei Liu .}
\thanks{Bingkun Nian, Jie Yang, and Wei Liu are with the School of Automation and Intelligent Sensing \& Institute of Image Processing and Pattern Recognition \& Institute of Medical Robotics, Shanghai Jiao Tong University, Shanghai 200240, China  (Email: \{sjtu\_nbk, jieyang, weiliucv\}@sjtu.edu.cn)}
\thanks{Fenghe Tang and Shaohua Kevin Zhou are with School of Biomedical Engineering, Division of Life Sciences and Medicine, University of Science and Technology of China, Hefei, Anhui, 230026, P.R. China. Suzhou Institute for Advanced Research, University of Science and Technology of China, Suzhou, Jiangsu, 215123, P.R. China. Center for Medical Imaging, Robotics, and Analytic Computing \& Learning (MIRACLE), Suzhou Institute for Advanced Research, USTC, Suzhou, P.R. China. (Email: 
fhtan9@mail.ustc.edu.cn; s.kevin.zhou@ustc.edu.cn)}
\thanks{Zhonglong Zheng is with the School of Computer Science and Technology, Zhejiang Normal University, Jinhua 321004, China (Email: zhonglong@zjnu.edu.cn).}
\thanks{Jianrui Ding is with the Harbin Institute of Technology, Heilongjiang, China (Email: jrding@hit.edu.cn).}
}

\markboth{Journal of \LaTeX\ Class Files,~Vol.~14, No.~8, August~2021}%
{Shell \MakeLowercase{\textit{et al.}}: A Sample Article Using IEEEtran.cls for IEEE Journals}


\maketitle

\begin{abstract}

Dynamic convolution demonstrates outstanding representation capabilities, which are crucial for natural image segmentation. However, it fails when applied to medical image segmentation (MIS) and infrared small target segmentation (IRSTS) due to limited data and limited fitting capacity. In this paper, we propose a new type of dynamic convolution called \textbf{d}ynamic \textbf{p}arameter \textbf{c}onvolution (DPConv) which shows superior fitting capacity, and it can efficiently leverage features from deep layers of encoder in reconstruction tasks to generate DPConv kernels that adapt to input variations.
% Moreover, we observe that DPConv, built upon deep features derived from reconstruction tasks, significantly enhances segmentation performance on the same dataset used for reconstruction. 
Moreover, we observe that DPConv, built upon deep features derived from reconstruction tasks, significantly enhances downstream segmentation performance. 
We refer to the segmentation network integrated with DPConv generated from reconstruction network as the \textbf{s}iamese \textbf{r}econstruction-\textbf{s}egmentation network (SRS). We conduct extensive experiments on seven datasets including five medical datasets and two infrared datasets, and the experimental results demonstrate that our method can show superior performance over several recently proposed methods. Furthermore, the zero-shot segmentation under unseen modality demonstrates the generalization of DPConv. The code is available at: https://github.com/fidshu/SRSNet.

\end{abstract}


% However, the above-mentioned networks focus solely on improving their own structures under segmentation tasks. Nevertheless, there exists a connection between different tasks, implying that we can combine segmentation tasks with reconstruction tasks to enhance network performance from a different perspective. This is what we will introduce: dynamic-parameter convolution and SRS.

\input{Section/introduction}

\input{Section/Relatework}

\input{Section/SRS}

\input{Section/Experiments}

\input{Section/Conclusion}

\bibliographystyle{unsrt}
\bibliography{main}

\input{Section/authors}

\end{document}

%% file: Section/introduction.tex
\section{Introduction}
\label{sec:intro}
% \IEEEPARstart{I}{n} the past decade, deep learning based methods have achieved significant success in various computer vision tasks including reconstruction \cite{SUPERRESOLUTION}, segmentation \cite{NNUNET}, etc. The most common approach to construct deep Convolutional Neural Networks involves stacking multiple convolutional layers and other fundamental layers organized with predefined feature connections. In segmentation tasks, the U-Net \cite{UNET} architecture, augmented with techniques such as Transformers and dynamic convolutions, has been widely applied. Meanwhile, the concept of dynamic convolutions is pioneered by Conditional parameterization (CondConv) \cite{CCONV}, Dynamic convolutions (DyConv) \cite{DYCONV}, Omni-Dimension dynamic convolution (ODConv) \cite{ODCONV}, and Spatial and Channel reconstruction convolution (ScConv) \cite{ScConv}. Typically, in contrast to regular convolutions that apply the same convolutional kernel to all input samples, dynamic convolutions learn linear combinations of different convolutional kernels. The attention weights for these linear combinations are dynamically computed based on input features. And with the help of dynamic convolutions, the accuracy of  CNNs has further improved.
\IEEEPARstart{O}{v}{e}{r} the past decade, deep learning-based methods have achieved remarkable success on weak target segmentation task, such as medical image segmentation (MIS)\cite{wang2022medical} and infrared small target segmentation (IRSTS) \cite{10960691}. 
Nonetheless, the gradual transitions in physiological tissue regions in medical images, coupled with the extreme sparsity of target features in infrared images, present significant challenges for further performance improvement of deep learning methods\cite{seg_application}. Dynamic convolutions provides a promising approach when applied to natural image segmentation. The concept of dynamic convolutions is pioneered by conditional convolution (CondConv) \cite{CCONV}, dynamic convolutions (DyConv) \cite{DYCONV}, omni-dimension dynamic convolution (ODConv) \cite{ODCONV}, spatial and channel reconstruction convolution (ScConv) \cite{ScConv} and dynamic on-demand convolution (DOD)\cite{xie2023learning}. Typically, unlike regular convolutions that apply a fixed convolutional kernel to all input samples, dynamic convolutions can learn linear combinations of multiple convolutional kernels\cite{DYCONV,ODCONV}. The attention weights for these linear combinations are dynamically computed based on the input features. And these approaches have led to further improvements in natural image\cite{shen2023adaptive,lei2023dgnet,wu2023dss}.

\begin{figure}[t]
    \centering
    \includegraphics[width = 0.4 \textwidth]{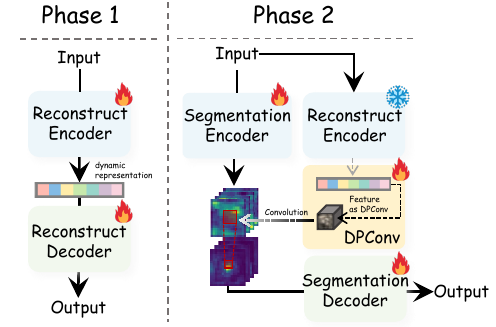}
    \caption{The proposed siamese reconstruction-segmentation network (SRS). Our method bridge the reconstruction and segmentation task through the proposed dynamic-parameter convolution (DPConv) which shows superior fitting capacity and can be additionally trained on the reconstruction task to boost the segmentation performance.}
    \label{fig:RSStructure}
\end{figure}

However, unfortunately, when dynamic convolution meet MIS or IRSTS, it faces the following two obvious challenges (discussed in Sec.~\ref{Subsection:PW}) : 1) \textbf{Limited labeled data}: Due to the factors such as disease mortality rates, ethical concerns surrounding medical data, or the military sensitivity of infrared data, these datasets are significantly more scarce than natural image datasets\cite{ssl3d, abd1k, word}. Furthermore, like conventional networks, traditional dynamic convolution networks require large labeled datasets for effective training\cite{ODCONV}. This presents considerable challenges for the application of dynamic convolution in MIS or IRSTS\cite{yang2025pinwheel}. 
2) \textbf{Limited fitting capacity}: 
% These methods involve the linear combination of different convolutional kernels, with the combination coefficients computed using various techniques. However, due to its mathematical properties, the convolutional kernels derived from a limited set still suffer from constrained fitting capacity. These limitations are particularly significant in weak target segmentation due to the exponential growth in computational complexity. 
These methods\cite{DYCONV,ODCONV} involve the linear combination of different convolutional kernels, with the combination coefficients computed using various techniques. However, due to its mathematical properties, the convolutional kernels derived from a limited set still suffer from limited fitting capacity. Moreover, the scarcity of medical data is insufficient to support adequate training of these sets of convolutional kernels\cite{tang2024hyspark}.
% The tissues in physiological regions of MIS \cite{said2022review} and extremely small targets in IRSTS \cite{nian2023local} also require accurate segmentations with low computational cost \cite{valanarasu2022unext,tong2024target}. This constraint further makes it impractical to stack large number of fixed convolutional kernels, as used in dynamic convolutions for natural images, thereby making them unsuitable for MIS and IRSTS.
The tissues in physiological regions of MIS \cite{said2022review} and extremely small targets in IRSTS \cite{nian2023local} also require accurate segmentations which is usually achieved by stacking larger numbers of fixed convolution kernels, however, these large numbers of kernels further require much more labeled training samples which are limited in MIS and IRSTS\cite{valanarasu2022unext,tong2024target}.

To address the aforementioned limitations, we propose a new type of dynamic convolution, called dynamic parameter convolution (DPConv). 
In MIS and IRSTS, it is difficult to obtain labeled images and access publicly available data sets tailored to specific tasks\cite{abd1k}. The high demand for large-scale datasets further complicates this issue, as extensive manual annotation by medical professionals is often impractical. However, collecting unlabeled images is comparatively feasible, presenting a valuable alternative\cite{mae}. 
Unlabeled images are typically employed in image reconstruction tasks ~\cite{bert,mae,simmim,jepa,cmae,spark} and we need to investigate how DPConv can be utilized to introduce the modality prior of the reconstruction task to the segmentation task, an area that has received limited attention in existing research.

The proposed DPConv is similar to mathematical convolution and consists of two components: DPConv kernel and operation convolution. DPConv kernel needs highly condensed semantic feature to dynamically generate the convolution kernel. Meanwhile the feature obtained from the deeper layers of the encoder of the reconstruction network can meet this requirement. At the same time, we use the generated DPConv kernel to convolve the feature passed from the segmentation network. It can naturally establish the relationship between reconstruction and segmentation tasks. This approach enables efficient utilization of large-scale datasets without relying heavily on labeled samples. When applied to segmentation tasks, the utilization of DPConv naturally forms a siamese structure, which we denote as the siamese reconstruction-segmentation network (SRS) (as illustrated in Fig.~\ref{fig:RSStructure}). Overall, the contributions of this paper are listed as follows:

% DPConv generates adaptive, dynamic-parameter kernels and the parameter generation component of DPConv relies on highly condensed semantic information, which is located in the deeper layers of the reconstruction network. 

% % A key rationale for selecting features in deep layers over shallow layers lies in the generator of DPConv. Using shallow features require a considerably larger generator consisted of multi layers, thus increasing complexity and computational cost. In contrast, deep layers features enable a lightweight generator architecture, effectively reducing both the parameter count and computational overhead.

% This approach enables efficient utilization of large-scale datasets without relying heavily on labeled samples. A complete DPConv consists of two main components: the encoder of reconstruction network and the dynamic-parameter kernel generation structure. Due to the unique design of DPConv, when applied to segmentation tasks, it naturally forms a siamese structure, which we refer to as the siamese reconstruction-segmentation (SRS) (as illustrated in Fig.~\ref{fig:RSStructure}). Overall, the contributions of this paper are listed as follows:
\begin{enumerate}
    \item  We propose a new type of dynamic convolution named DPConv. It can directly generate convolution kernel which is different from the existing dynamic convolutions that generate combination coefficients.

    % \item A novel siamese method, named SRS, is proposed based on the DPConv module. In this framework, the reconstruction task serves as the information source for DPConv, which is effectively applied to the segmentation task. We further demonstrates that SRS improves segmentation performance by leveraging additional training data from the reconstruction task, without the need to increase the training data specifically for segmentation.

    % \item A novel siamese method, named SRS, is first proposed to unify the reconstruction task and segmentation task by utilizing DPConv. The reconstruction task serves as the information source for DPConv, which is effectively applied to the segmentation task. By comprehensive experiments, we further demonstrates that SRS improves segmentation performance by leveraging additional training data from the reconstruction task, without the need to increase the training data specifically for segmentation. 

    \item A novel siamese method, named SRS, is first proposed to bridge the reconstruction task and segmentation task by utilizing DPConv. The reconstruction task serves as the information source for the parameter generation of DPConv kernel which is then effectively applied to the segmentation task. In addition, our SRS can simply improve the segmentation performance  by leveraging additional training data from the reconstruction task, without the need to increase the training data specifically for segmentation. 

    \item We validate our method on five MIS datasets and two IRSTS datasets. Comprehensive experimental results demonstrate that our method can achieve superior performance over the state-of-the-art approaches.
\end{enumerate}

The rest of this paper is organized as follows: Sec.~\ref{Sec:RelatedWork} is devoted to the related work. We introduce our DPConv and SRS in Sec.~\ref{Sec:network}.
We compare our method with several state-of-the-art approaches and evaluate its effectiveness on multiple medical and infrared datasets in Sec.~\ref{Sec:ExperimentSetting}.
We draw the conclusion of this paper in  Sec.~\ref{Sec:Conclusion}.

%% file: Section/Relatework.tex
\section{Relate Work}
\label{Sec:RelatedWork}
\subsection{Dynamic Convolution}
Dynamic convolution plays a crucial role in natural image segmentation\cite{wang2023internimage}. CondConv \cite{CCONV}, one of the most well-known dynamic convolution techniques, operates as a linear combination of multiple static convolutions, in which the convolution kernels are weighted based on the input. This makes CondConv input-dependent, enabling it to adapt the convolutional process according to the specific characteristics of the input. CondConv utilizes a single attention scalar value, meaning that all output filters share the same attention value for a given input. It neglects the spatial dimensions of the convolutional kernels as well as the input and output channel dimensions, resulting in a limited and coarse exploration of the kernel space. This makes the CondConv show relatively modest performance gains in larger networks\cite{CCONV}. ODConv\cite{ODCONV} addresses these limitations by introducing a multi-dimensional attention mechanism using a parallel strategy, enabling the learning of complementary attention across four dimensions of the kernel space. Different from previous dynamic convolution, Pinwheel\cite{yang2025pinwheel} changes the convolution range from matrix to wind turbine, while FDConv\cite{chen2025frequency} completes convolution from the frequency perspective.
However, due to the gradual transitions in physiological tissue regions in MIS and the extreme sparsity of targets in IRSTS, a more refined exploration of the kernel structure is necessary for more accurate segmentation.
Moreover, another limitation of existing dynamic convolution techniques in MIS and IRSTS is the scarcity and high cost of labeled data. Traditional dynamic convolution\cite{ODCONV,CCONV, DYCONV}, like standard convolution, is still constrained by the availability of labeled data for effective training. In contrast, unlabeled data is more readily obtainable~\cite{sslmia, mim, survey}, making it essential to develop strategies that can fully exploit these data, and our method is properly designed to achieve this goal.

\subsection{Weak Target Segmentation}
Weak target segmentation (WTS), which mainly includes MIS and IRSTS, differs significantly from natural image segmentation. The primary distinctions lie in the presence of extremely sparse semantic features and gradual transitions in the target regions, such as tissue regions in MIS\cite{SWINUNET,REPVIT} and sky or ocean regions in IRSTS\cite{DNANET}. To improve segmentation accuracy in WTS, deep learning methods, exemplified by ERDUNet~\cite{ERDUnet}, are actively exploring novel architectures to addresss the aforementioned challenges. ERDUnet achieves significant advances by introducing an encoder module capable of simultaneously extracting local features and capturing global continuity information. 
% Similar to ERDUnet, various methods have been proposed \cite{POLYYP, NOVEL, LIUDEEP, WANGWIDE} to enhance the performance of WTS. 
Currently, hybrid models combining Transformers\cite{MOBILEVIT,TRANSFORMER} and convolutional neural networks (CNNs) are widely applied in tasks such as multi-modal brain tumor segmentation \cite{TRANSBTS} and 3D image segmentation \cite{COTR}. 

% However, the aforementioned networks primarily focus on optimizing their internal architectures for segmentation tasks in isolation. However, there is a potential connection between different tasks, suggesting that integrating segmentation tasks with reconstruction tasks could provide a novel approach to enhancing network performance. In this context, we introduce DPConv and the SRS to leverage this relationship.

However, the aforementioned networks primarily focus on optimizing their internal architectures for segmentation tasks in isolation. However, there is a potential connection between different tasks like DRSL\cite{wang2020dual}, which employs a three-stage framework: first, a pre-trained segmentation task supervises the image super-resolution task, then the image super-resolution task utilizes a similarity matrix to fine-tune the segmentation decoder. In contrast, our approach adopts a two-stage framework that leverages reconstruction tasks to supervise DPConv generation for segmentation. Fundamentally, DRSL represents a fine-tuning methodology, while our SRS provides a dynamic convolution generation paradigm.

% In the field of weak image segmentation including MIS and IRSTS, many innovations regarding network structures have been proposed\cite{JIENCODER}. 
% Recently, The ERDUnet~\cite{ERDUnet} achieves nice improvements. It improves the feature extraction efficiency by constructing an encoder module that can simultaneously extract local features and global continuity information. some novel attention modules are designed to optimize segmentation boundary regions while improving training efficiency. The feature transfer structure of the decoding part is also improved, which fully integrates the features of different levels to restore the spatial resolution more finely.
% Similar to ERDUnet\cite{ERDUnet}, many methods attemps \cite{POLYYP,NOVEL,LIUDEEP,WANGWIDE} to improve the performance of medical segmentation. Presently, various combinations of Transformers and CNNs are applied in multi-modal brain tumor segmentation~\cite{TRANSBTS} and 3D segmentation~\cite{COTR}. Furthermore, pure Transformers have demonstrated remarkable potentials in medical image segmentation~\cite{SWINUNET}.

%% file: Section/SRS.tex
\section{Siamese Reconstruction-Segmentation Network Based on Dynamic-Parameter Convolution}
\label{Sec:network}
In order to improve the performance of segmentation networks, we propose the dynamic parameter convolution and further show the connection between the low-level reconstruction task and the high-level segmentation task.

\subsection{Overview of the Network Architecture}

\begin{figure*}[t]
    \centering
    \includegraphics[width = 0.8 \textwidth]{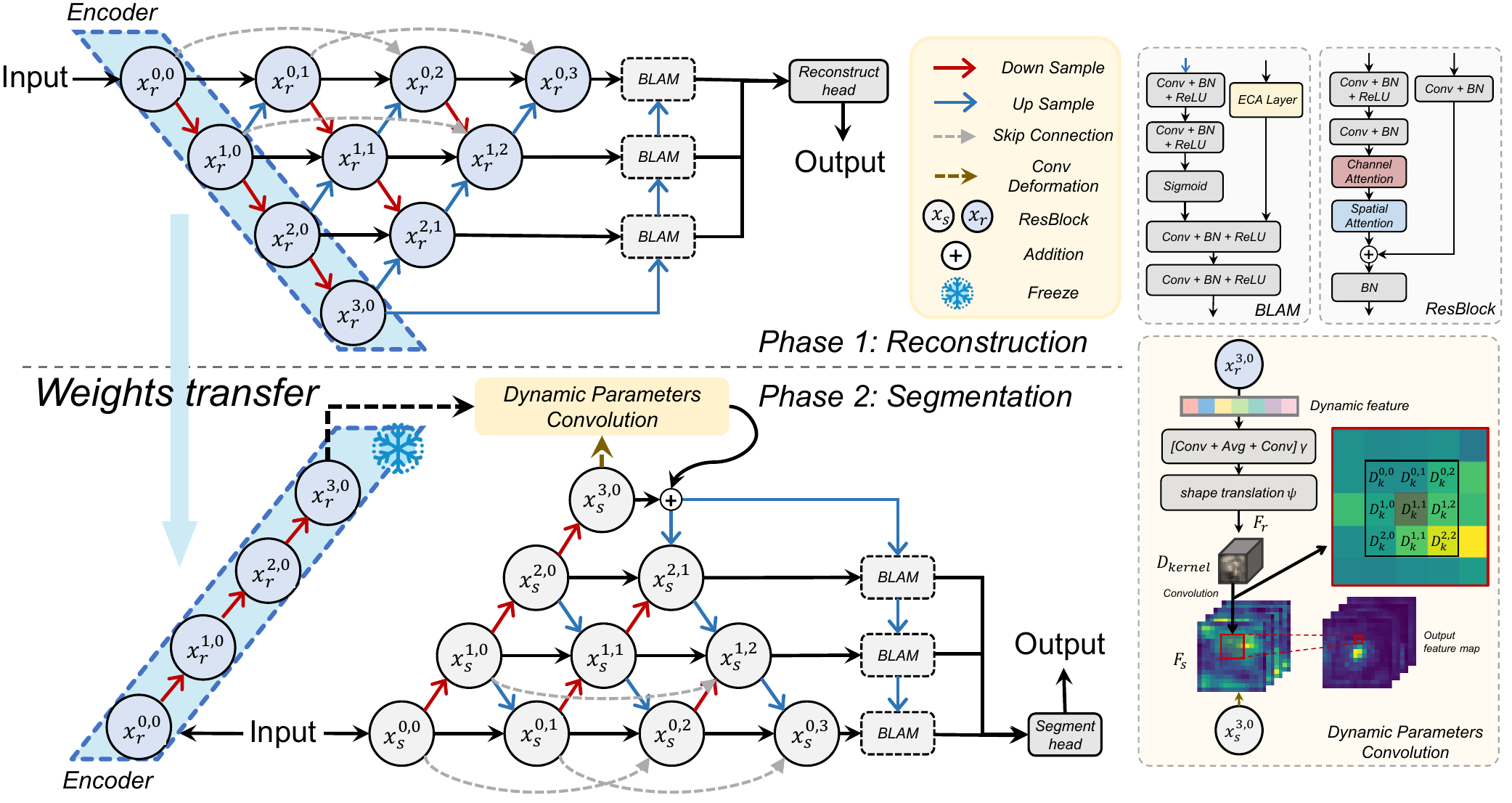}
    \caption{
    % The schematic diagram of the SRSNet illustrates a unified structure consisting of two tasks: reconstruction and segmentation. The upper part of the network is dedicated to the reconstruction task, while the lower part is designed for the segmentation task. The connection between two tasks is facilitated by the DPConv.
    The structure of our SRS consists of two networks: reconstruction and segmentation. Resblock $x_{r},x_{s}$ are the convolution block $x$ of the reconstruction network and the convolution block of the segmentation network, respectively. In the phase one, the network is trained on the reconstruction task. In the phase two, the trained reconstruction network is frozen and the encoder component of the reconstruction network is extracted for segmentation training. The feature $F_{r}$ extracted from $x_{r}^{3,0}$ generates the DPConv kernel $D_{kernel}$. The DPConv kernel is then convolved with the feature $F_{s}$ passed from $x_{s}^{3,0}$. 
    }
    \label{fig:SRSnetwork_STRUCTURE}
\end{figure*}

The overall structure of the siamese reconstruction-segmentation network (SRS) is illustrated in Fig.~\ref{fig:SRSnetwork_STRUCTURE}. The upper part is dedicated to the reconstruction task, while the lower part is designed for the segmentation task. DPConv serves as the bridge between these two networks.
It contains two components: dynamic parameter convolution (DPConv) kernel and convolution operation. The DPConv kernel is generated from the deep layers of the reconstruction encoder and convolved with the feature passed from the segmentation encoder. The network contains ResBlock~\cite{RESNET} and BLAM block\cite{ALCNET}.

% Now, the key to establishing a connection between the compression reconstruction task and the segmentation task lies in the DPConv. Thereforce, In Section.\ref{Subsection:DPC}, we will offer an detailed explanation of the structure and construction methods employed in DPConv. Then in Section.\ref{Subsection:RSS}, a detailed description of how to construct the SRSnetwork will be provided.

\subsection{Dynamic Convolution Revisit}
\label{Subsection:PW}
In order to adapt to various changes in data distribution, many methods related to dynamic convolutions have been proposed such as CConv~\cite{CCONV}, Dyconv~\cite{DYCONV} and OD-Conv~\cite{ODCONV}. Their mathematical descriptions can be summarized as follows:
\begin{equation}
\label{ClassicDyConvMath}
    \omega = \sum_{i=1}^{n} \sum_{j=1}^{n} \alpha_{ji}\omega_{i},
\end{equation}
where $\omega$, $\alpha_{ji}$, $\omega_{i}$ are the new kernel, combination coefficient and the fixed sets of convolutional kernels.

Both DyConv~\cite{DYCONV} and OD-Conv~\cite{ODCONV} employ a fixed number of kernels, denoted as $\omega_{i}$. The primary distinction between them lies in the number of coefficient sets utilized: DyConv generates a single set, whereas OD-Conv produces four sets.

{Although DyConv and OD-Conv replace a single fixed convolution with multiple convolutions, they fundamentally rely on a finite set of fixed convolutional kernels once the training of the network is completed. 
% This implies that the dynamic parameters are unable to adapt based on the input and the weights can only adapt to a fixed number of situations.
This implies that when training ends, the weight of the fundamental kernel $\omega_{i}$ is fixed. The fixed $\omega_{i}$ cannot be efficiently adapted to the medical image because the lesions can exhibit different forms of biological tissue. 
Moreover, the constantly increasing quantity of fixed kernels is difficult for medical image and infrared image. The scarcity of medical data is insufficient to support adequate training of these sets of convolutional kernels.}

\subsection{Dynamic Parameter Convolution}
\label{Subsection:DPC}

\begin{algorithm}
\caption{{DPConv generation subnetwork $\gamma$ }}
\label{alg:p_kernel_d2}
\begin{algorithmic}[1]
\Require 
    {$x_1$}: {Input feature map} {$\in \mathbb{R}^{B \times C_{in} \times H \times W}$} \\
    {$\mathit{weight}$: {Informative feature map} $\in \mathbb{R}^{B \times C_{info} \times H' \times W'}$}
\Ensure 
    {$\mathit{out}$: {Output feature map} $\in \mathbb{R}^{B \times C_{out} \times H \times W}$}
    
\Statex
\Function{{Forward}}{{$x_1, \mathit{weight}$}}
    \State ${\mathit{kernel}} \gets \text{{ConvNet}}{(\mathit{weight})}$  
    \Comment{{Dynamic kernel generation}}
    \State ${\mathit{kernel}} \gets \text{{reshape}}(\mathit{kernel}, B \times C_{out}, C_{in}, K, K)$
    \Comment{{Kernel dimension transformation}}
    \State $x \gets \text{{reshape}}(x_1, 1, B \cdot C_{in}, H, W)$  
    \Comment{{Input feature reshape} $\psi$}
    \State $\mathit{out} \gets \text{{conv2d}}\big(x, \mathit{kernel}, \text{{pad}}=\lfloor (K-1)/2 \rfloor, \text{{stride}}=1, \text{{groups}}=B\big)$ 
    \State $\mathit{out} \gets \text{{reshape}}(\mathit{out}, B, C_{out}, H, W)$  
    \Comment{{Output dimension restoration}}
    \State \Return $\mathit{out}$
\EndFunction
\Statex
\Function{{Initialization}}{}
    \State $C_{in} \gets \text{{input channels (default: 8)}}$
    \State $C_{out} \gets \text{{output channels (default: 1)}}$
    \State $K \gets \text{{kernel size (default: 1)}}$
    \State $C_{info} \gets \text{{information channels (default: 64)}}$
    \State $\mathcal{H} \gets C_{in} \times C_{out} \times K^2$  
    \Comment{{Hidden dimension specification}}
    \State $\text{{ConvNet}} \gets \left[
        \begin{array}{l}
            \text{{Conv2d}}(C_{info}, C_{info}, 1) \\
            \text{{AdaptiveAvgPool2d}}(1) \\
            \text{{GroupConv}}(C_{info}, \mathcal{H}, 1)
        \end{array}
    \right]$  
    \Comment{{Kernel generation network}}
\EndFunction
\end{algorithmic}
\end{algorithm}

To solve the problem mentioned above, we propose DPConv, which is shown in Fig.~\ref{fig:SRSnetwork_STRUCTURE}. 
As shown in Alg.~\ref{alg:p_kernel_d2}, DPConv adopts a compact learnable network that directly generates the parameters of convolutional kernels, which is totally different from the conventional approaches \cite{DYCONV,ODCONV,ScConv} that generate the combination coefficients for fixed convolutional kernels.
Once the training is completed, the proposed compact network can dynamically generate kernel parameters based on the input. 

In our DPConv, the feature maps from the encoder of the reconstruction network serve as input to generate DPConv kernel. However the selection between shallow and deep features plays a crucial role in determining the efficiency of the compact network. 
The generated DPConv kernel is used for segmentation. It is necessary to offer the condensed semantic features \cite{xie2023learning,ODCONV} which are usually located at the deep layer of the encoder. In addition, shallow features require a larger generator with multiple layers, significantly increasing both complexity and computational cost. In contrast, using deep feature maps enables a more lightweight generator architecture, effectively reducing both the number of parameters and the computational overhead.

% \begin{figure}[t]
%     \centering
%     \includegraphics[width = 0.5 \textwidth]{NetworkStructure/Dconv1.png}
%     \caption{The DPConv structure comprises two components. $\mathcal{E}_{r}$ and $\mathcal{E}_{s}$ are the encoder of the reconstruction and segmentation networks. $\mathcal{D}_{s}$ is the decoder of segmentation network. The upper structure provides information for generating dynamic-parameter kernel  component and the lower provides input data convoluted.}
%     \label{fig:DYNAMICPARAMETERCONVLUTION}
% \end{figure}

% After presenting the schematic diagram of DPConv, we also provide its mathematical description. Given that the information in the segmentation task is denoted as $X$, and the information in the reconstruction task is denoted as $Y$, with a transformation operation $\psi$, the dynamic-parameter convolution process can be described by the following formula:
We first generate DPConv kernel: denote the channel number of the input features as $M$, the kernel size as $K \times K$ and the desired channel number of the output features as $N$, and then the total number of parameters that need to be generated is $n = M \times N \times K^{2}$. Thus, in order to get $n$ parameters, we chose to generate a vector $\theta_{n}$ from a compact network $\gamma$ consisting of convolution, avg-pooling and convolution. Given the reconstruction condensed semantic feature $F_{r}$ from the last layer of the encoder of the reconstruction network, shape translation $\psi$ that translates $\theta_n$ into a 2D convolution kernel, the DPConv kernel $D_{kernel}$ can be described as:
% \begin{equation}
%     D_{kernel}(F_{r}) = \psi(\theta_{n}) = \psi(\gamma(F_{r})).
% \end{equation}

% \begin{align}
%     & D_{kernel}(F_{r}) = \psi(\theta_{n}), \\
%     & \psi(\theta_{n}) = \psi(\gamma(F_{r})).
% \end{align}
\begin{align}
    & \theta_{n} = \gamma(F_{r}), \\
    & D_{kernel}(F_{r}) = \psi(\theta_{n})
\end{align}
After obtaining DPConv kernel $D_{kernel}$, we can provide the whole running process of DPConv. Given that the semantic feature $F_{s}$ from the last layer of the segmentation encoder and the $F_{r}$ from the last layer of the reconstruction encoder, the mathematical description of DPConv can be described as follows:
\begin{equation}
\label{Operation:DPConv}
    DPConv(F_{s},F_{r}) = Conv2d(F_{s},D_{kernel}(F_{r})),
\end{equation}
where Eq.~\ref{Operation:DPConv} denotes convoluting $F_s$ with $D_{kernel}(F_r)$, and in this process, each convolutional kernel parameter is dynamically generated.

% To perform the DPConv operation, the convolutional kernel must possess specific parameters: the desired number of target channels is represented as $M$, the kernel size is indicated by $K$, and the output channels are denoted as $N$. The total number of parameters to generate can be expressed as $n = M \times N \times K^{2}$.  Thus, in order to generate $n$ parameters, we choose to construct a one-dimensional vector $\theta_{n}$ with size $n$.

% To get $\theta_{n}$, the semantic feature $Y \in R^{z=C \times H \times W}$ needs to be translated to one-dimensional vector $\theta_{n}$ . If we directly convert $Y$ into $\theta_{n}$ using average operations, it will lead to non-convergence of the network because of small receptive field. Therefore, it is necessary to use multiple convolutions and downsampling for achieving one-dimensional vector $\theta_{n}$, which is effective in increasing the receptive field and preventing overfitting. Given reconstruction information input $Y$, the semantic translation $\psi_{2}$ can be described as: 
% \begin{equation}
% \label{EQ:ChanelTrans}
%     \psi_{2}(Y) = \sigma_{kernel=1}^{z \Rightarrow n}(\delta(\sigma_{kernel=1}^{z \Rightarrow z}(Y)))
% \end{equation}
% In Eq.~\ref{EQ:ChanelTrans}, we first employ one $1 \times 1$ convolution $\sigma_{kernel=1}^{z \Rightarrow z}$ and adaptive AvgPool $\delta$ to adjust the channel dimensionality. Then, we use another $1 \times 1$ convolution $\sigma_{kernel=1}^{z \Rightarrow n}$ to reshape feature from $Y$ to $\theta_{n}$ and change channels from $z$ to $n$.

\subsection{Siamese Reconstruction-Segmentation Network}
\label{Subsection:RSS}

% Now, We have presented the structure of DPConv in Fig.~\ref{fig:DYNAMICPARAMETERCONVLUTION} . The key question of constructing network comes out: \textit{how to establish a relationship between two tasks (reconstruction and segmentation) using DPConv ?} 
The features in the deeper layer have better semantic information and the existing dynamic convolutions such as DOD\cite{xie2023learning}, ODConv\cite{ODCONV} often use the feature in the last layer of the encoder as input to generate a dynamic kernel. These condensed semantic features from the last layer can help to efficiently improve the segmentation accuracy. However, due to the scarcity of medical datasets and infrared datasets, the DPConv kernel generator may not be fully trained. 
As a result, we use the feature in the last layer of the encoder of the reconstruction network as the input for DPConv kernel generator. This is because the features in the last layer of the encoder of the reconstruction network can also contain condensed semantic information while the reconstruction task does not require labeled samples.
In addition, there can be much more training data available for reconstruction task than segmentation task.
We denote this novel structure, consisting of the reconstruction network, DPConv and segmentation network, as siamese reconstruction-segmentation network (SRS).

% In order to solve this challenging question, we utilize DPConv proposed before as the connection between the low-level reconstruction task and the high-level segmentation task (shown in Fig.~\ref{fig:SRSnetwork_STRUCTURE}). DPConv can forms a novel training structure named siamese reconstruction-segmentation, in which the upper and lower parts representing dual tasks. 

The SRS (shown in Fig.~\ref{fig:SRSnetwork_STRUCTURE}) is divided into two primary stages: the first phase involves the reconstruction network, while the second phase focuses on the segmentation network. The reconstruction and segmentation network follows a U-shaped architecture, but with modules that are carefully designed. The encoder comprises Resblock block and downsampling while the decoder consists of cascade upsampling resblock and dense connections for skip connection.

1) \textbf{Phase one (reconstruction)}: The reconstruction task provides highly condensed information, which is crucial for generating the parameters of DPConv kernel. Therefore, constructing DPConv kernel requires initial training on a reconstruction task. Upon completion of reconstruction training, the deep features can be considered as highly compressed semantic representations. These compressed features greatly aid in the construction of DPConv kernel, allowing it to efficiently capture suitable patterns according to the input feature. 
% The reconstruction loss $\mathcal{L}_{r}$ is L2 loss and defined as follows:
The $L2$ loss between the reconstruct image and the input is adopted to train the reconstruction network, defined as follows:
\begin{equation}
    \mathcal{L}_{r} = \frac{1}{2N}\sum_{i=1}^N (\Tilde{R_{i}} - R_{i})^2
\end{equation}
where $\Tilde{R_{i}}$ is the reconstructed image generated by the decoder of the reconstruction network and $R_{i}$ is the input image.

2) \textbf{Phase two (segmentation)}: During the phase two, the encoder of the reconstruction network is frozen, while the segmentation task is normally trained. As noted previously, the reconstruction network adopts an encoder-decoder architecture. However, DPConv utilizes only the feature in the last layer of the encoder, the decoder and densenested connection is unnecessary. Thus, these components are removed, and only the encoder is retained and frozen. The segmentation loss $\mathcal{L}_{s}$ is defined as a combination of the binary cross entropy loss (BCE) and dice loss (Dice) between the predicted result $\Tilde{S_{i}}$ and the ground truth $S_{i}$:
\begin{equation}
    \mathcal{L}_{s}  = 0.5 \times BCE(\Tilde{S_{i}},S_{i}) + DICE(\Tilde{S_{i}},S_{i})
\end{equation}

In the aforementioned two stages, we have incorporated both DPConv and the reconstruction module. Their additional resource consumption, including parameters (Params), GFLOPs, and runtime FPS, is presented in Table~\ref{tab:consumption}. The reconstruction module introduces minimal overhead, with only 482K additional parameters and 1.615 GFLOPs, while maintaining real-time performance at 25 FPS. DPConv causes even smaller overhead, adding only 0.006 GFLOPs. When integrated into the segmentation task, the SRS structure, which combines DPConv and the reconstruction module, results in negligible increases in Params, GFLOPs, and FPS compared to the baseline segmentation network.

% To balance the additional computational load and performance improvements introduced by DPConv, we adjusted the number of stacked modules, differentiating between the reconstruction network and the segmentation network. The specific network configuration parameters are outlined in Table~\ref{tab:DifferentSettings}.

Overall, the advantages of the SRS are listed as follows: On the one hand, the reconstruction network in phase one allows to gather a substantial amount of unlabeled images as training data. This enables us to create a reconstruction network that possesses robust adaptability and is not constrained to specific tasks. On the other hand, with the help of DPConv, the performance of the segmentation network within the phase two can improve without altering the structure framework, or training strategy of the network.

\begin{figure*}[!t]
\centering
\subfloat[BUS]{\includegraphics[width=0.135\textwidth]{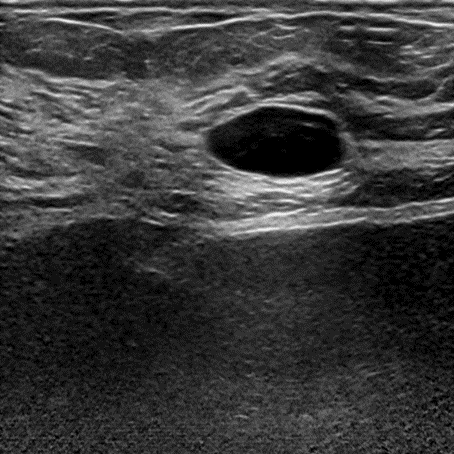}}
\hfill
\subfloat[BUSI]{\includegraphics[width=0.135\textwidth]{NetworkStructure/images/busi.png}}
\hfill
\subfloat[TNSCUI]{\includegraphics[width=0.135\textwidth]{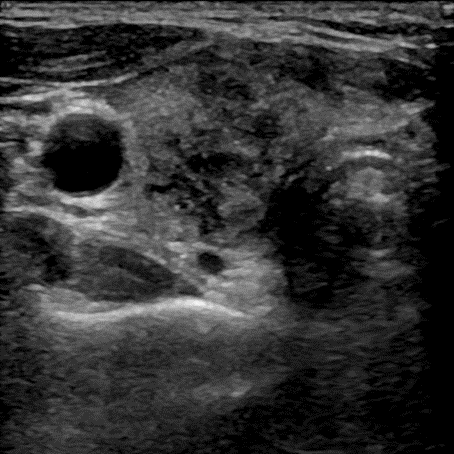}}
\hfill
\subfloat[ISIC 2018]{\includegraphics[width=0.135\textwidth]{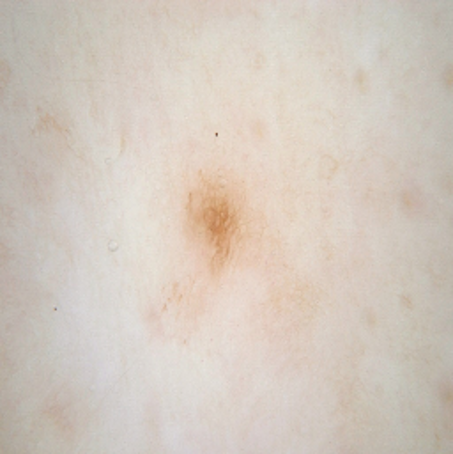}}
\hfill
\subfloat[Synapse]{\includegraphics[width=0.135\textwidth]{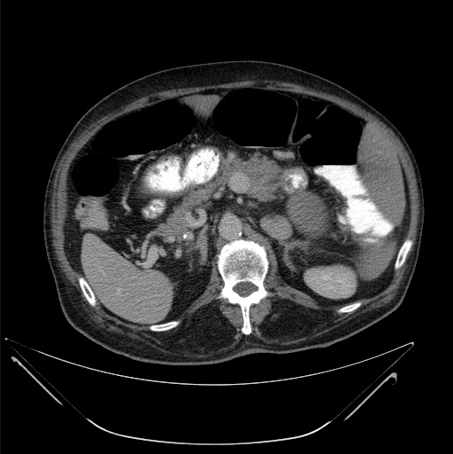}}
\hfill
\subfloat[NUDT-KBT19]{\includegraphics[width=0.135\textwidth]{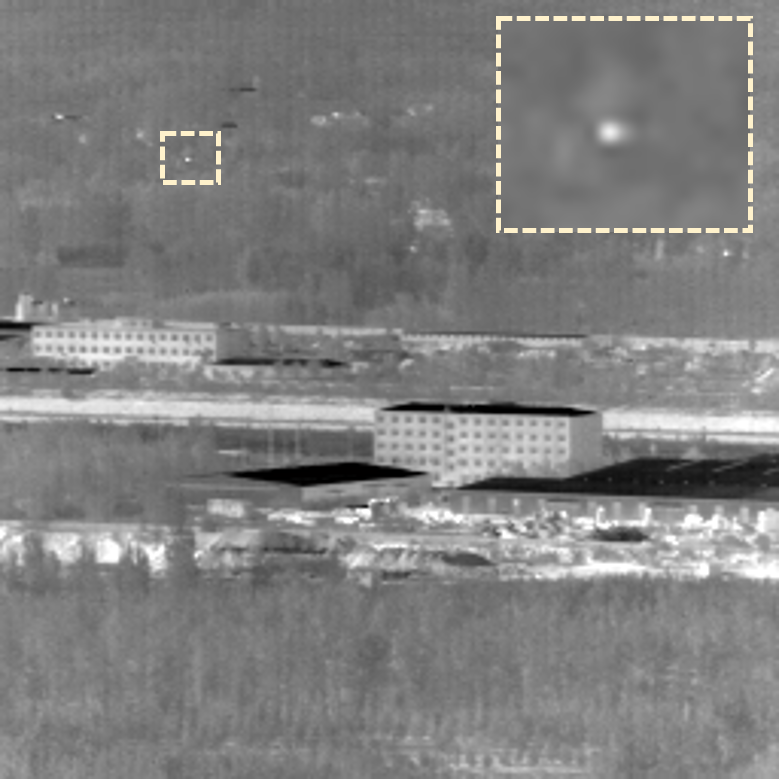}}
\hfill
\subfloat[SIRST]{\includegraphics[width=0.135\textwidth]{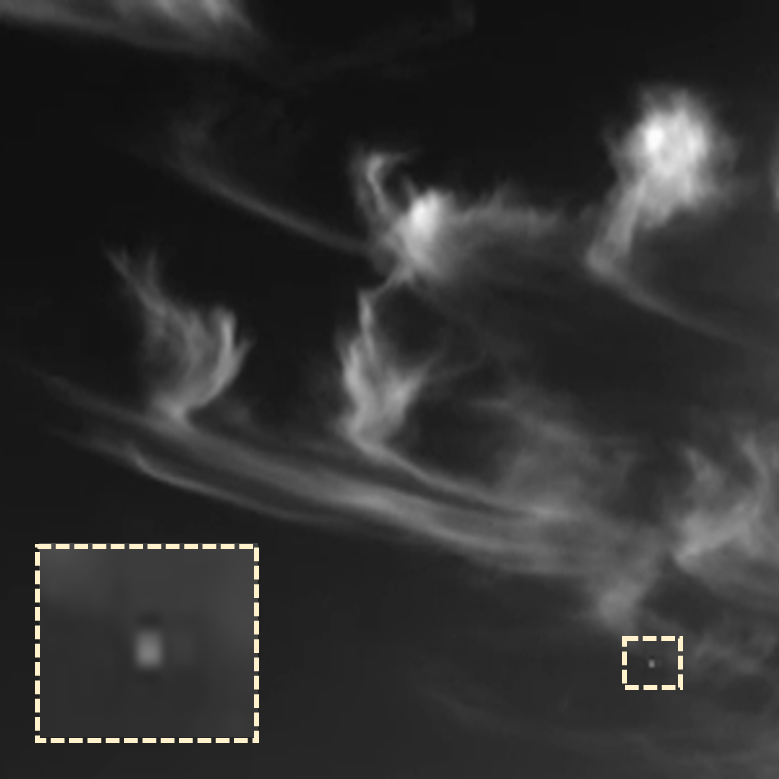}}
\caption{Images from different datasets. The targets to be segmented basically do not contain valid internal information, and the remaining valid information is also less different from external clutters.}
\label{fig:WeakTargetImage}
\end{figure*}

% \begin{table}[!]
% \caption{Settings for reconstruction and segmentation networks\label{tab:DifferentSettings}}
% \resizebox{1\linewidth}{!}
% {
% \begin{tabular}{c | c |c c   }
% \hline % horizontal line

% \multirow{2}{*}{Network}
% & \multirow{2}{*}{Quantity} & \multicolumn{2}{|c}{Training Parameters} \\
% \cline{3-4}
% && Ratio & Additional  \\
% \hline
% Reconstruction            & [5,5,5,5]  & 100\% & Yes \\
% Segmentation       & [6,8,12,6] & 70\%   & No \\
% \hline
% \end{tabular}
% }

% \end{table}

% \begin{table*}[!]
% \centering
% \caption{Consumption of DPConv\label{tab:consumption}}
% \resizebox{0.75\linewidth}{!}
% {
% \begin{tabular}{c | c c c | c |c c   }
% \hline % horizontal line

% \multirow{2}{*}{Network} & \multicolumn{3}{c|}{Module}
% & \multirow{2}{*}{Params} & \multicolumn{2}{c}{Computation} \\
% \cline{2-4}
% & Reconstruction& DPConv & Segmentation  && GFLOPS & FPS  \\
% \hline
% Pure Reconstruction          &\ding{52}&&& 482K  & 1.617G & 25.72 \\
% DPConv                  &&\ding{52}&& 1.18M  & 0.006G & - \\
% Pure Segmentation            &&&\ding{52}& 9.17M & 47.461G & 15.01 \\ 
% SRS            &\ding{52}&\ding{52}&\ding{52}&  10.35M & 47.467G  & 14.17 \\
% \hline
% \end{tabular}
% }

% \end{table*}

\begin{table}[!t]
\centering
\caption{Performance Analysis of DPConv Components}
\label{tab:consumption}
\renewcommand{\arraystretch}{1.2}
\footnotesize
\begin{tabularx}{\columnwidth}{c|c|c|c|r|r|r}
\toprule
\multirow{2}{*}{{Network Config.}} & \multicolumn{3}{c|}{{Components}} & \multirow{2}{*}{{Params}} & \multicolumn{2}{c}{{Computation}} \\
\cmidrule(lr){2-4} \cmidrule(l){6-7}
& {R.} & {DPC} & {S.} & & {GFLOPs} & {FPS} \\
\midrule
Pure Reconstruction & \checkmark & -- & -- & 482K & 1.617 & 25.72 \\
DPConv Only & -- & \checkmark & -- & 1.18M & 0.006 & -- \\
Pure Segmentation & -- & -- & \checkmark & 9.17M & 47.461 & 15.01 \\
SRS & \checkmark & \checkmark & \checkmark & 10.35M & 47.467 & 14.17 \\
\bottomrule
\end{tabularx}
\end{table}

%% file: Section/Experiments.tex
\section{Experiments}
\label{Sec:ExperimentSetting}
\subsection{Dataset}
As shown in Fig.~\ref{fig:WeakTargetImage}, the WTS datasets comprise two main categories, namely medical datasets and infrared datasets. In medical image dataset, three modalities medical images are selected: CT, ultrasound and dermoscopy image. In infrared datasets, small target image dataset is selected. Each dataset is divided into 7:3 for training data and test data.

\noindent\textbf{Synapse Dataset}. Synapse multi-organ segmentation dataset\footnote{\url{https://www.synapse.org}} comprises abdominal CT scans of 8 organs from 30 cases (3779 axial images). Each CT volume consists of $85 \sim 198$ slices of 512$\times$512 pixels, with a voxel spatial resolution of ($[0.54 \sim 0.54]\times[0.98 \sim 0.98]\times[2.5 \sim 5.0]$) mm$^3$. The dataset is divided into 2654 images for training and 1134 images for test.

\noindent\textbf{BUS Dataset}~\cite{DATA_BUS}. The Breast UltraSound (BUS) dataset\footnote{\url{http://cvprip.cs.usu.edu/busbench/}} contains 562 breast ultrasound images collected using five different ultrasound devices, including 306 benign cases and 256 malignant cases, each with the corresponding ground-truth segmentation mask. 
The dataset is divided into 393 images for training and 169 images for test.

\noindent\textbf{BUSI Dataset}~\cite{DATA_BUSI}. The Breast UltraSound Images (BUSI) dataset\footnote{\url{https://scholar.cu.edu.eg/?q=afahmy/pages/dataset}} collected from 600 female patients, includes 780 breast ultrasound images, covering 133 normal cases, 487 benign cases, and 210 malignant cases, each with corresponding ground truth labels. Following recent studies~\cite{UNEXT, CMUNET}, we only utilize benign and malignant cases from this dataset. 
The dataset is divided into 546 images for training and 234 images for test.

\noindent\textbf{TNSCUI Dataset}\cite{DATA_TNSCUI}. The Thyroid Nodule Segmentation and Classification in Ultrasound Images 2020 (TNSCUI) dataset\footnote{\url{https://tn-scui2020.grand-challenge.org/Dataset/}} collected from different ages and genders, includes 3644 cases, covering thyroid nodule, each with corresponding ground truth labels. 
The dataset is divided into 2550 images for training and 1094 images for test.

\noindent\textbf{ISIC 2018 Dataset}. The International Skin Imaging Collaboration (ISIC 2018) dataset\footnote{\url{https://challenge.isic-archive.com}} contains 2594 dermoscopic lesion segmentation images. 
The dataset is divided into 1815 images for training and 779 images for test. 

\noindent\textbf{SIRST Dataset}\cite{DATA_SIRST}. 
SIRST\footnote{\url{https://github.com/yiMISndai/sirst/tree/master}} is a dataset containing 427 images specially constructed for single-frame infrared small target detection, in which the images are selected from hundreds of infrared sequences for different scenarios.
The dataset is divided into 298 images for training and 129 images for test.

\noindent\textbf{NUDT-KBT19 Dataset}~\cite{DATA_NUDT-KBT19}. The 2019 Aerospace Cup infrared dataset\footnote{\url{http://www.csdata.org/p/387/}} contains 5993 infrared images with weak and small targets like unmanned aerial vehicles collected under different meteorological conditions. 
The dataset is divided into 4194 images for training and 1799 images for test. 

\subsection{Zero-shot Dataset}

\noindent\textbf{TUCC}. {The Thyroid Ultrasound Cine-clip (TUCC) dataset\footnote{\url{https://aimi.stanford.edu/datasets/thyroid-ultrasound-cine-clip}} is collected from 167 patients with biopsy-confirmed thyroid nodules (n=192) at the Stanford University Medical Center. The dataset consists of ultrasound cine-clip images, radiologist-annotated segmentations, patient demographics, lesion size and location, TI-RADS descriptors, and histopathological diagnoses. We test its 17K video frames.}

\noindent\textbf{PH2}. {The PH2 dataset~\cite{ph2} contains 200 dermoscopic images. The manual segmentation, clinical diagnosis, and identification of many dermoscopic structures, carried out by skilled dermatologists, are included in the PH2.}

\noindent\textbf{IRSTD-1K}. {IRSTD-1K\cite{zhang2022isnet} is a real-world infrared small target detection dataset containing 1,001 manually annotated real images with various target shapes, different target sizes, and rich clutter backgrounds from diverse scenarios. The target proportion in the entire image is approximately 0.019\%.}

\subsection{Implement Details}
The reconstruction network is trained first. During the training of the reconstruction network, the network is trained independently on each single dataset. Once the reconstruction network training is complete, we will freeze the encoder of the reconstruction network and train the segmentation network with DPConv.

We resize all the training images to 256$\times$256 and apply random rotation and flip for data augmentations. In addition, we use the SGD optimizer with a weight decay of $1e^{-4}$ and a momentum of 0.9 to train the networks. The initial learning rate is set to 0.01 and the decay is set as follows:
\begin{equation}
    decay = (1.0 - E_{\alpha} / E_{\text{MaxE}})^{0.9},
\end{equation}
where $\alpha$, $E_{\text{MaxE}}$ means the current epoch and the max epoch, respectively.
The batch size of reconstruction network is set to 8 with the network trained for 100 epochs. The batch size of segmentation network is set to 8 with the network trained for 300 epochs. All the experiments are conducted on the NVIDIA GeForce RTX 2080Ti GPU.

\begin{table*}[!t]
\centering
\caption{Quantitative evaluation of the results on the ultrasound and dermoscopy datasets. \gray{Grey} indicates datasets not mentioned in the original paper. Best results are highlighted as \colorbox{l}{\bf \!first\!}, \colorbox{ll}{\!second\!}.}
\label{tab:ExOnUlt_Der} 
\resizebox{\textwidth}{!}
{
\begin{tabular}{l l |cc|cccccc|cc|cc}
\hline
\multirow{3}{*}{Methods} & \multirow{3}{*}{\# Publication} & \multicolumn{2}{c|}{Computation} & \multicolumn{6}{c|}{Ultrasound} & \multicolumn{2}{c|}{Dermoscopy} & \multicolumn{2}{c}{CT} \\
\cline{3-14}
& &\multirow{2}{*}{Params} & \multirow{2}{*}{GFLOPs} & \multicolumn{2}{c}{BUS} & \multicolumn{2}{c}{BUSI} & \multicolumn{2}{c|}{TNSCUI}& \multicolumn{2}{c|}{ISIC} & \multicolumn{2}{c}{Synapse} \\
% \cline{5-18}
&&& & IoU & F1 & IoU & F1 & IoU & F1 & IoU & F1 & Dice & HD95 \\
\hline
\multicolumn{14}{c}{\textbf{Specialized Models (One model for one medical task)}} \\
\hline

% \textit{{CNN Network}} &&&&&&&&& \\
\hline
{DRSL\cite{wang2020dual}} & CVPR'20&0.99 &1.19 & \gray{83.42} & \gray{91.78} & \gray{65.25} & \gray{74.73} & \gray{81.23} & \gray{88.52} & \gray{70.21} & \gray{80.03} & \gray{60.62} & \gray{42.58} \\
% U-Net~\cite{UNET} & MICCAI'15 & 34.52 & 65.52   & 86.73 & 92.46 & 68.61 & 76.97 & 75.88 & 84.24 & 82.18 & \underline{89.97} & 78.35 & 85.91 \\

MedT~\cite{MEDT}  & MICCAI'21 & 1.37 & 2.40  & \gray{80.81} & \gray{88.78} & \gray{63.36} & \gray{73.37} & \gray{71.00} & \gray{80.87} & \gray{81.79} & \gray{89.74} & \gray{55.21} & \gray{60.06}\\   

UNeXt~\cite{UNEXT}  & MICCAI'22 & 1.47 & 0.58 & \gray{84.73} & \gray{91.20} & 65.04 & 74.16 & \gray{71.04} & \gray{80.46} & 82.10 & 89.93 & \gray{69.99} & \gray{41.43}\\

SwinUnet~\cite{SWINUNET} & ECCV'22 & 27.14 & 5.14  & \gray{85.27} & \gray{91.99} & \gray{63.59} & \gray{76.94}  & \gray{75.77} & \gray{85.82} & \gray{82.15} & \gray{89.98} & 74.13 & 28.54 \\

MissFormer~\cite{MISSFORMER} & TMI'22 & 35.45 & 7.25  & \gray{81.92} & \gray{89.46} & \gray{63.29} & \gray{73.47} & \gray{68.26} & \gray{76.71} & \gray{80.99} & \gray{88.03} & 72.53 & 29.79\\

EGE-Unet~\cite{EGEUNET} & MICCAI'23 & 0.045 & 0.072   & \gray{84.72} & \gray{91.72} & \gray{58.90} & \gray{74.11} & \gray{74.47} & \gray{85.36} & 82.19 & \cellcolor{l}{\textbf{90.22}} & \gray{60.38} & \gray{57.12}\\

CMU-Net~\cite{CMUNET}  & ISBI'23 & 49.93 & 91.25  & \gray{87.18} & \gray{92.89} & \cellcolor{ll}{71.42} & 79.49 & \gray{77.12} & \gray{85.35} & 82.16 & 89.92 & \gray{74.49} & \gray{36.76}\\

ERDUNet\cite{ERDUnet} & TCSVT'23 & 10.21 & 10.29  & \gray{85.41} & \gray{91.64} & \gray{68.67} & \gray{77.74} & \gray{76.02} & \gray{84.63} & 82.03 & 88.96 & \gray{68.60} & \gray{44.40}\\

MCDANet~\cite{MCDANet}  & SPL'24 & 48.75 & 58.58 & \gray{86.93} & \gray{92.82} & \gray{70.23} & \gray{79.22} & \gray{77.84} & \gray{86.13} & \gray{82.07} & \gray{89.49} & \gray{-} & \gray{-}\\

{UCTransNet~\cite{yuan2024sctransnet}} &AAAI'22 & 66.24 & 32.98 & \gray{85.98} & \gray{92.03} & \gray{70.05} & \gray{78.51} & \gray{75.51} & \gray{84.08} & \gray{82.76} & \gray{89.50} & {77.82} & {33.60}\\

{TinyU-Net~\cite{chen2024tinyu}} & MICCAI'24 & 0.48 & 1.66 & \gray{85.67} & \gray{91.86} & \gray{66.21} & \gray{75.01} & \gray{74.03} & \gray{82.95} & {81.95} & {88.99} & \gray{74.45} & \gray{43.19} \\

{EMCAD\cite{rahman2024emcad}} & CVPR'24 & 26.76 & 5.60 & \cellcolor{ll}{87.81} & \cellcolor{ll}{93.30}  & \gray{71.09} & \cellcolor{ll}{81.40} & \cellcolor{ll}{77.48} & \cellcolor{ll}{85.83} & \cellcolor{ll}{82.91} & \gray{89.56} &  \cellcolor{l}{\textbf{81.82}} & \cellcolor{l}{\textbf{19.94}}\\

% {MADGNet\cite{nam2024modality}} & CVPR'24 & 31.56 & 10.08 & 88.58 & 94.27 & 72.92 & 81.04 & 79.02 & 86.92 & 83.72 & 90.21 & 79.28 & 29.74 \\
\hline
\multicolumn{14}{c}{\textbf{Generation Models (One model for two  modality)}} \\
\hline
MobileViT-xxs~\cite{MOBILEVIT} & ICLR'22 & 4.30 & 0.55 & \gray{81.62} & \gray{89.42} & \gray{63.00} & \gray{73.71} & \gray{70.16} & \gray{80.40} & \gray{79.76} & \gray{87.65} & \gray{52.58} & \gray{31.35}\\  

MobileViT-xs~\cite{MOBILEVIT} & & 5.77 & 1.16 & \gray{82.13} & \gray{89.69} & \gray{63.63} & \gray{74.19} & \gray{71.13} & \gray{81.19} & \gray{79.91} & \gray{87.73} & \gray{54.73} & \gray{31.02}\\

MobileViT-s~\cite{MOBILEVIT} & & 10.66 & 2.13 & \gray{82.57} & \gray{89.99} & \gray{64.28} & \gray{74.68} & \gray{71.64} & \gray{81.60} & \gray{80.12} & \gray{87.89} & \gray{55.57} & \gray{30.81}\\

EdgeViT-xxs~\cite{EDGEVITS} & ECCV'22 & 6.83 & 0.90 & \gray{81.30} & \gray{89.12} & \gray{61.36} & \gray{72.12} & \gray{68.36} & \gray{78.82} & \gray{78.92} & \gray{86.97} & \gray{49.59} & \gray{36.97}\\  

EdgeViT-xs~\cite{EDGEVITS} & & 10.15 & 1.70 & \gray{81.86} & \gray{89.55} & \gray{61.94} & \gray{72.58} & \gray{69.18} & \gray{79.44} & \gray{79.17} & \gray{87.15} & \gray{51.80} & \gray{28.71}\\

EdgeViT-s~\cite{EDGEVITS} & & 16.49 & 2.71 & \gray{81.32} & \gray{89.13} & \gray{61.12} & \gray{71.79} & \gray{68.74} & \gray{79.19} & \gray{79.06} & \gray{87.06} & \gray{50.68} & \gray{31.82}\\

EMO-1m~\cite{EMOVIT} & ICCV'23 & 3.36 & 0.54 & \gray{84.68} & \gray{91.44} & \gray{67.06} & \gray{77.11} & \gray{73.01} & \gray{82.76} & \gray{80.97} & \gray{88.46} & \gray{59.35} & \gray{32.15}\\
EMO-2m~\cite{EMOVIT} & & 4.58 & 0.87 & \gray{85.11} & \gray{91.73} & \gray{67.90} & \gray{77.87} & \gray{73.91} & \gray{83.39} & \gray{81.33} & \gray{88.77} & \gray{60.60} & \gray{32.47}\\
EMO-5m~\cite{EMOVIT} & & 7.87 & 1.70 & \gray{85.06} & \gray{91.66} & \gray{68.27} & \gray{78.29} & \gray{74.34} & \gray{83.77} & \gray{81.50} & \gray{88.90} & \gray{59.08} & \gray{25.00}\\
EMO-6m~\cite{EMOVIT} & & 9.14 & 1.85 & \gray{84.89} & \gray{91.58} & \gray{67.50} & \gray{77.71} & \gray{74.02} & \gray{83.53} & \gray{81.31} & \gray{88.74} & \gray{59.47} & \gray{27.68}\\ 

RepViT-m1~\cite{REPVIT} & CVPR'24 & 8.48 & 1.34 & \gray{76.73} & \gray{85.98} & \gray{55.52} & \gray{67.18} & \gray{66.61} & \gray{77.48} & \gray{78.15} & \gray{86.43} & \gray{49.00} & \gray{32.71}\\
RepViT-m2~\cite{REPVIT} & & 12.49 & 2.09 & \gray{75.93} & \gray{85.37} & \gray{54.70} & \gray{66.27} & \gray{64.88} & \gray{75.99} & \gray{78.03} & \gray{86.34} & \gray{47.91} & \gray{32.26}\\
RepViT-m3~\cite{REPVIT} & & 14.17 & 1.85 & \gray{76.51} & \gray{85.79} & \gray{56.07} & \gray{67.62} & \gray{66.21} & \gray{77.09} & \gray{78.34} & \gray{86.62} & \gray{47.61} & \gray{30.69}\\
\hline
\textbf{SRS} &     & 10.35 & 47.46       & \cellcolor{l} \textbf{88.29} & \cellcolor{l}\textbf{93.44} & \cellcolor{l}\textbf{74.47} & \cellcolor{l}\textbf{82.57} & \cellcolor{l}\textbf{79.33} & \cellcolor{l}\textbf{87.02} & \cellcolor{l}\textbf{83.16} & \cellcolor{ll} 89.72 & \cellcolor{ll} 79.80 & \cellcolor{ll} 24.51\\
\hline
\end{tabular}
}
\end{table*}

\subsection{Evaluation Metrics and Training Strategy}
We primarily utilize three widely recognized evaluation metrics to validate the performance of our approach. These metrics include Intersection over Union (IoU) and F1 score for the 2D datasets: BUS, BUSI, TNSCUI, ISIC2018, SIRST, NUDT-KBT19 datasets; Hausdorff Distance (HD95), and the Dice for the 3D datasets: Synapse.

{For medical imaging datasets (BUS, BUSI, TNSCUI, and Synapse), which are provided without predefined train/val/test splits, we perform three randomized 7:3 splits (70\% training, 30\% testing) to ensure statistical robustness. The ISIC2018 dataset is similarly processed with three randomized 7:3 splits, consistent with the methodology employed by UNeXt, CMU-Net, and other approaches. The same splitting strategy is applied to infrared datasets SIRST (7:3) and NUDT-KBT19 (7:3), ensuring methodological uniformity across all benchmarks. To mitigate randomness, each model is retrained three times per dataset, and the final results are derived from the mean performance. Regarding the reconstruction task dataset scale, as shown in Table~\ref{tab:reconstructionmethod}, "Resp." represents the use of segmentation datasets only, while "Joint" denotes the incorporation of external datasets. Additionally, for Synapse 3D data, all models operate exclusively on 2D slices which is the same as \cite{SWINUNET}.}

\begin{figure*}[!ht]
  \centering
   \includegraphics[width=0.8\linewidth]{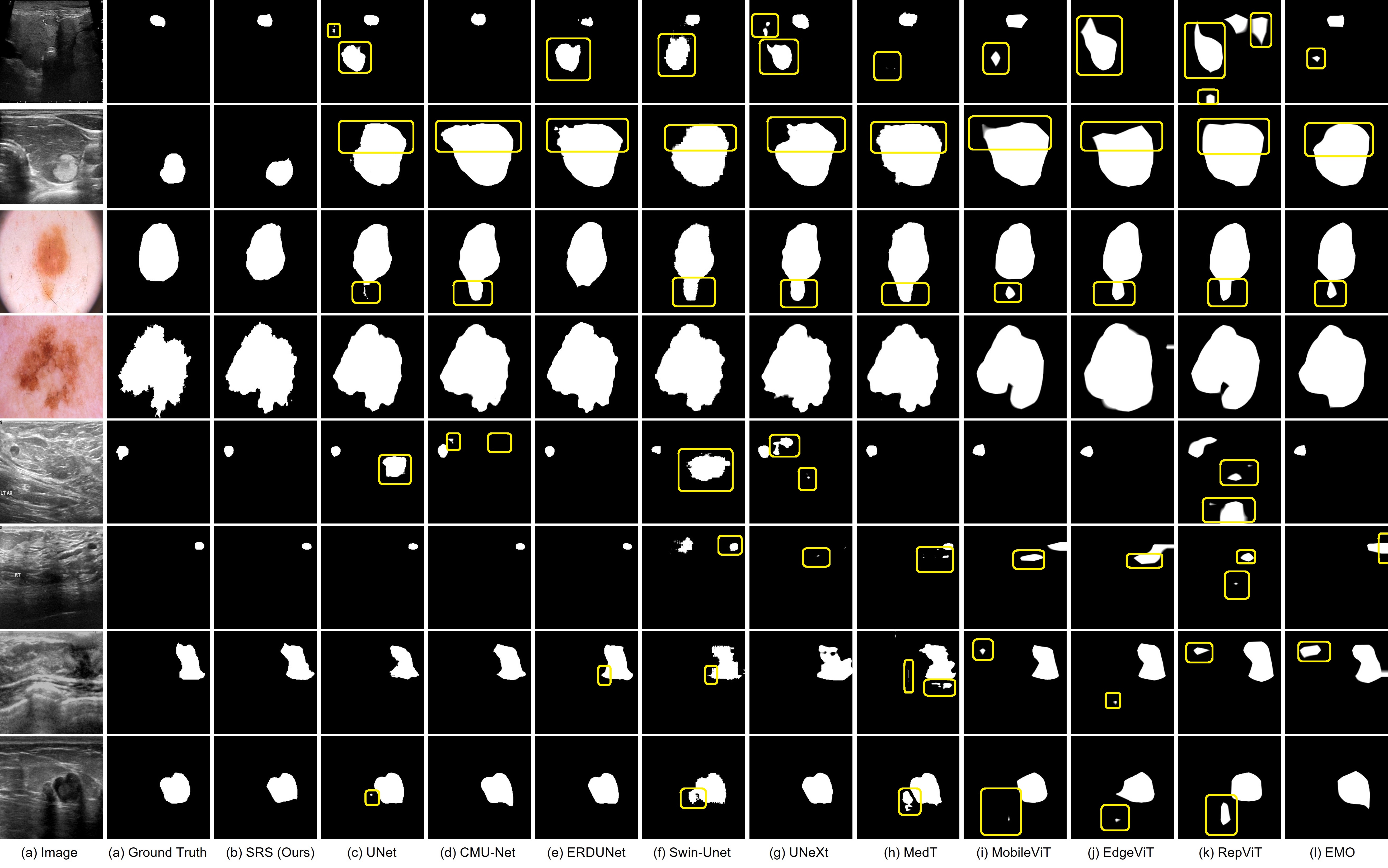}
   \caption{Visual comparison of the results produced by UNet~\cite{UNET}, CMU-Net~\cite{CMUNET}, ERDUNet~\cite{ERDUnet}, Swin-Unet~\cite{SWINUNET}, UNeXt~\cite{UNEXT}, MedT~\cite{MEDT}, MobileViT~\cite{MOBILEVIT}, EdgeViT~\cite{EDGEVITS}, RepViT~\cite{REPVIT} and EMO~\cite{EMOVIT} on Ultrasound and Dermoscopy Dataset. 
   % Row 1 and 2 - TNSCUI samples, Row 3 and 4 - ISIC18 samples, Row 5 and 6 – BUSI samples, Row 7 and 8 - BUS samples. Yellow boxes represent error segmentation.
   Images in rows 1-2 are from the TNSCUI dataset, 
   images in rows 3-4 are from the ISIC18 dataset,
   images in rows 5-6 are from the BUSI dataset,
   images in rows 7-8 are from the BUS dataset.
   Yellow boxes represent error segmentations.
   }
    \label{fig:fourds}
\end{figure*}

% \begin{figure*}[!ht]
%   \centering
%    \includegraphics[width=0.8\linewidth]{NetworkStructure/label_image/supply_ct.png}
%    \caption{Visual comparison of results produced by Swin-Unet~\cite{SWINUNET}, UNeXt~\cite{UNEXT}, MobileViT~\cite{MOBILEVIT}, EdgeViT~\cite{EDGEVITS}, RepViT~\cite{REPVIT}, EMO~\cite{EMOVIT} and MedT~\cite{MEDT} on Synapse dataset.}
%    \label{fig:supply_ct}
% \end{figure*}

\begin{table}[t]
\centering
\caption{Experiments on infrared small target datasets. \gray{Grey} indicates datasets not mentioned in the original paper. Best results are highlighted as \colorbox{l}{\bf \!first\!}, \colorbox{ll}{\!second\!}. \label{tab:ExOnISTDataset}}
\resizebox{0.5\textwidth}{!}{%
\begin{tabular}{l l |cc|cccc}
\hline
\multirow{3}{*}{Methods} & \multirow{3}{*}{\# Publication} & \multicolumn{2}{c|}{Computation} & \multicolumn{4}{c}{Infrared target} \\
\cline{3-8}
&&\multirow{2}{*}{Params} & \multirow{2}{*}{GFLOPs} & \multicolumn{2}{c}{SIRST} & \multicolumn{2}{c}{NUDT-KBT} \\
&&& & IoU & F1 & IoU & F1 \\
\hline
\multicolumn{8}{c}{\textbf{Generation Models (One model for two modality)}} \\
\hline
MobileViT-xxs~\cite{MOBILEVIT} & ICLR'22 & 4.30 & 0.55 &\gray{69.37} &\gray{80.67} &\gray{84.49} &\gray{90.43}\\  
MobileViT-xs~\cite{MOBILEVIT} & & 5.77 & 1.16 &\gray{68.78} &\gray{80.42} &\gray{84.86} &\gray{90.60}\\
MobileViT-s~\cite{MOBILEVIT} & & 10.66 & 2.13 &\gray{69.44} &\gray{80.76} &\gray{84.85} &\gray{90.68}\\
EdgeViT-xxs~\cite{EDGEVITS} & ECCV'22 & 6.83 & 0.90 &\gray{62.83} &\gray{75.25} &\gray{75.74} &\gray{81.62}\\  
EdgeViT-xs~\cite{EDGEVITS} & & 10.15 & 1.70 &\gray{61.59} &\gray{74.20} &\gray{75.96} &\gray{81.64}\\
EdgeViT-s~\cite{EDGEVITS} & & 16.49 & 2.71 &\gray{62.30} &\gray{74.78} &\gray{83.77} &\gray{89.87}\\
EMO-1m~\cite{EMOVIT} & ICCV'23 & 3.36 & 0.54 &\gray{67.06} &\gray{78.74} &\gray{85.41} &\gray{91.02}\\
EMO-2m~\cite{EMOVIT} & & 4.58 & 0.87 &\gray{67.47} &\gray{79.21} &\gray{85.24} &\gray{90.92}\\
EMO-5m~\cite{EMOVIT} & & 7.87 & 1.70 &\gray{66.83} &\gray{78.79} &\gray{85.06} &\gray{90.78}\\
EMO-6m~\cite{EMOVIT} & & 9.14 & 1.85 &\gray{67.06} &\gray{78.94} &\gray{85.48} &\gray{90.97}\\ 
RepViT-m1~\cite{REPVIT} & CVPR'24 & 8.48 & 1.34 &\gray{62.58} &\gray{74.84} &\gray{84.13} &\gray{90.18}\\
RepViT-m2~\cite{REPVIT} & & 12.49 & 2.09 &\gray{62.89} &\gray{75.11} &\gray{84.16} &\gray{90.16}\\
RepViT-m3~\cite{REPVIT} & & 14.17 & 1.85 &\gray{63.53} &\gray{75.91} &\gray{84.64} &\gray{90.51}\\
\hline

\multicolumn{8}{c}{\textbf{Specialized Models (One model for one infrared small task)}} \\
\hline
AGPCNet~\cite{AGPCNET} & TAES'23 & 12.36 &43.18 &\gray{69.25} &\gray{80.54} &\gray{84.88} &\gray{90.80} \\
ACM~\cite{ACM} & WACV'21 &8.27 &7.87 &{72.10} &{81.42} &\gray{77.16} &\gray{86.46} \\
ALCNet~\cite{ALCNET} & TGRS'21& 0.37 &3.69&{72.80} &{83.67} &\gray{80.22} &\gray{87.66} \\
DNANet~\cite{DNANET} & TIP'23&4.69 &14.27&{73.13} &{83.87} &\gray{85.66} &\gray{91.13} \\
RPCANet\cite{wu2024rpcanet} & WACV'24&12.36 &43.18 &{69.02} &{79.78} &\gray{81.00} &\gray{87.95} \\
ID-UNet\cite{chen2025id} & AEJ'25& 16.93 &1.14&{70.64} &{81.66} &\cellcolor{ll}{86.14} &\cellcolor{ll}{91.57} \\
{MSHNet\cite{liu2024infrared}} & CVPR'24& 4.06 &6.11 &{72.10} &{83.43} &\gray{85.14} &\gray{90.84} \\
{SCTransNet\cite{yuan2024sctransnet}} & TGRS'24&11.19 &20.24 &{73.03} &\cellcolor{ll}{84.49} &- &- \\
{Pinwheel\cite{yang2025pinwheel}} & AAAI'25&2.54 &4.60 &\cellcolor{ll}{73.23} &{83.87} &\gray{85.01} &\gray{90.77} \\
{FDConv\cite{chen2025frequency}} & CVPR'25&0.458 &0.393 &{56.63} &{69.36} &\gray{83.12} &\gray{89.44} \\
\hline
\textbf{SRS} & & 10.35 &47.467 & \cellcolor{l}{\textbf{74.67}} & \cellcolor{l}{\textbf{84.28}} & \cellcolor{l}{\textbf{86.34}} & \cellcolor{l}{\textbf{92.62}}\\
\hline
\end{tabular}%
}
\end{table}

\subsection{Experiments on Medical Images}
\label{Sec:Experiment}
Fig.~\ref{fig:fourds} presents the results of various networks on medical datasets. The visual representation clearly demonstrates that the proposed SRS based on DPConv surpasses other state-of-the-art networks in terms of segmentation quality. Detailed quantitative evaluations are presented in the following subsections for an in-depth analysis of the proposed method.

\subsubsection{Experiments on Ultrasound and Dermoscopy Datasets}

% In ultrasound (BUS, BUSI, TNSCUI) and dermoscopy (ISIC2018) datasets, our proposed SRS is compared with other state-of-the-art networks in Tab.~\ref{tab:ExOnUlt_Der}. Quantitative results show that our SRS achieves the best performance in both two kinds of datasets.

% The BUS and BUSI datasets are used for ultrasound image segmentation of breast cancer lesions. On both datasets, SRS demonstrates superior performance in terms of the primary evaluation metric, IoU. Specifically, on the BUS dataset, SRS achieves an improvement of 1.11\%, reaching 88.29\% compared to 87.18\% with other methods. On the BUSI dataset, the improvement is even more pronounced, with SRS surpassing competing methods by 3.05\%, achieving 74.47\% compared to 71.42\%. Similarly, on the TNSCUI dataset, which is used for thyroid lesion segmentation, SRS outperforms other networks, achieving an improvement of 2.21\% (79.33\% vs. 77.12\%).

% The ISIC 2018 dataset is a skin cancer segmentation dataset captured under natural lighting conditions. We observe that networks that perform suboptimally on ultrasound images, such as EGE-Unet and MissFormer, achieve relatively strong results on this dataset. The primary reason is that the skin cancer dataset contains rich details, including texture, contrast, and well-defined edges. The distribution of the test data aligns closely with that of the training data, which simplifies the training process and enables all networks to perform effectively in this scenario.

In 2D medical segmentation including ultrasound (BUS, BUSI, TNSCUI) and dermoscopy (ISIC2018) datasets, our proposed SRS is compared with other state-of-the-art networks in Tab.~\ref{tab:ExOnUlt_Der}. Quantitative results show that our SRS achieves the best performance in both types of data set. Specifically, We observe that networks that perform suboptimally on ultrasound images, such as EGE-Unet and MissFormer, achieve relatively strong results on this dataset. The primary reason is that the skin cancer dataset contains rich details, including texture, contrast, and well-defined edges. The distribution of the test data aligns closely with that of the training data, which simplifies the training process and enables all networks to perform effectively in this scenario.

Additionally, the aforementioned experiments show another advantage of SRS: its ability to achieve the best performance across datasets captured under different reaction medium. This advantage, indicative of the generalizability of the network, is also demonstrated in Tab.~\ref{tab:ExOnUlt_Der} and Tab.~\ref{tab:zeroshot}.

\begin{figure}[!]
    \centering
    \includegraphics[width=0.9\linewidth]{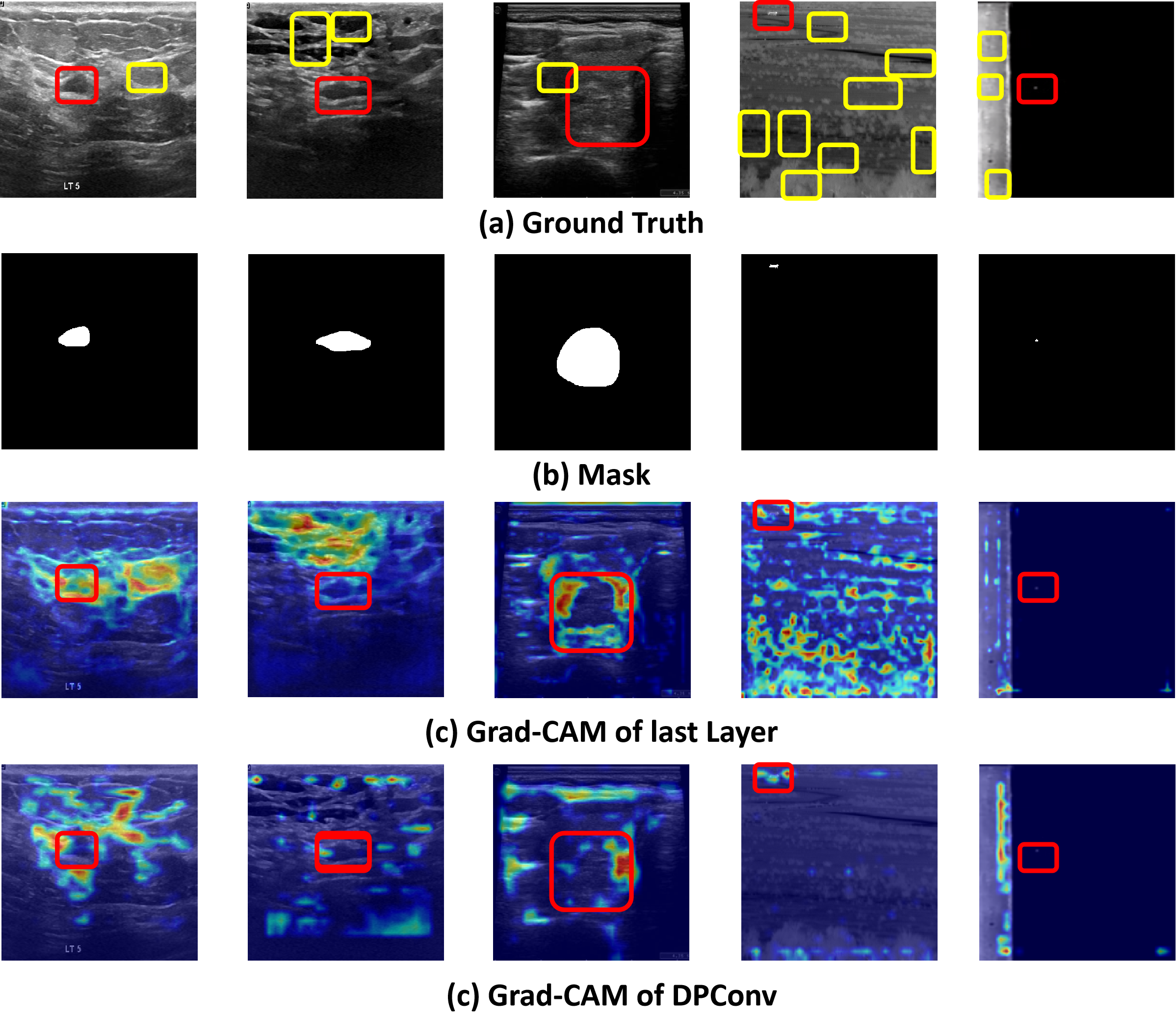}
    \caption{{Visualization of Grad-CAM activations corresponding to the encoder outputs. Red box and yellow box means correct and failure attention area, respectively.}}
    \label{fig:failure_case}
\end{figure}

\begin{table*}[t]
\centering
\caption{{Generalization experiment of different model zero-shot on the unseen dataset. Source training dataset $\rightarrow$ zero-shot test dataset. Best results are highlighted as} \colorbox{l}{\bf \!first\!}, \colorbox{ll}{\!second\!}.}

\resizebox{0.8\textwidth}{!}
{%
\begin{tabular}{l|l|cc cc|cc|cc}
\hline
\multirow{3}{*}{Networks}&\multirow{3}{*}{Publication} & \multicolumn{4}{c|}{Ultrasound} & \multicolumn{2}{c|}{Dermoscopy} & \multicolumn{2}{c}{Infrared image} \\
\cline{3-10}
&& \multicolumn{2}{c}{BUSI $\rightarrow$ BUS} & \multicolumn{2}{c|}{TNSCUI $\rightarrow$ TUCC} & \multicolumn{2}{c|}{ISIC $\rightarrow$ PH2} & \multicolumn{2}{c}{SIRST $\rightarrow$ IRSTD-1K}\\
\cline{3-10}
&& IoU(\%) & F1(\%) & IoU(\%) & F1(\%) & IoU(\%) & F1(\%) & IoU(\%) & FD(\%)\\
\hline
UNeXt &MICCAI'22 & {69.02} & {79.27} & {54.83} & {66.04} & {82.00} & {89.57} &- & -\\
MissFormer & TMI'22 & \cellcolor{ll}{72.81} & \cellcolor{ll}{82.27} & \cellcolor{l}{\textbf{83.53}} & \cellcolor{l}{\textbf{90.15}} & {69.82} & {78.02} &- & - \\
ERDNet &TCSVT'23 & {71.75} & {80.72} & {59.66} & {70.06} & {82.90} & {90.06} &- & -\\
RepViT-m3 &CVPR'24 & {65.07} & {76.38} & {57.28} & {68.88} & {79.95} & {88.27} &- & -\\
MobileViT-s &CVPR'23 & {70.70} & {81.14} & {59.83} & {70.78} & \cellcolor{ll}{83.19} & \cellcolor{ll}{90.42} &- & -\\
TinyU-Net & MICCAI'24 & {64.49} & {73.70} & {57.76} & {69.03} & {81.96} & {89.53} &- & -\\
SCTransunet & TGRS’24 & - & - & - & - & - & - & {46.97} & {84.01} \\
MSHNet & CVPR’24 & - & - & - & - & - & - & {46.95} & {84.14} \\
PinWheel & AAAI’25 & - & - & - & - & - & - & \cellcolor{ll}{53.81} & \cellcolor{l}{\textbf{88.77}} \\
FDConv & CVPR’25 & - & - & - & - & - & - & {0.025} & {1.7} \\
\hline
SRS & ours & \cellcolor{l}{\textbf{74.21}} & \cellcolor{l}{\textbf{83.16}} & \cellcolor{ll}{60.58} & \cellcolor{ll}{70.83} & \cellcolor{l}{\textbf{85.33}} & \cellcolor{l}{\textbf{91.66}} & \cellcolor{l}{\textbf{54.36}} & \cellcolor{ll}{86.05} \\
\hline
\end{tabular}
}
\label{tab:zeroshot}
\end{table*}

\subsubsection{Experiments on Computed Tomography Images}
% In our comprehensive evaluation, we also perform experiments on CT images (3D segmentation). The networks selected are different than 2D segmentation task (ultrasound, Dermoscopy). The evaluation metrics employ for CT images are mIoU, Dice, and HD95. The experimental results are summarized in Tab.~\ref{tab:ExOnMRI}. Notably, apart from the versatile SwinUnet and our SRS, both transformer-based networks and  cnn-based networks experienced a significant decline in performance. For instance, compared to SRS, models such as MedT, EdgeViT, and MobileViT, among others, exhibited performance drops of up to 20\%. Similarly, cnn-based networks also demonstrated a performance decrease of approximately 10\%.

{We further test our method on the 3D segmentation for the CT images. We handle 3D data by slicing them into 2D slices, which are then fed into the network for segmentation.} The evaluation metrics employed for the segmentation results of CT images are Dice and HD95. Tab.~\ref{tab:ExOnUlt_Der} shows the quantitative evaluation of the results produced by different approaches. In particular, apart from EMCAD and our SRS, other methods suffer from a significant decline in performance. 
For instance, compared to SRS, the performance of models in 3D segmentation such as MedT, EdgeViT, MobileViT, MobileViT, RepViT, EMO, UNeXt, ERDUNet and UniRepLKNet exhibit a larger accuracy margin, ranging from 10\% - 20\%. These findings highlight the superior performance of our proposed structure, SRS, in CT tasks. Notably, SRS attains the second highest accuracy while maintaining a lower parameter.
%################1：目前现象是什么(详细写，结合数据定量说明一下好在哪几个方面)###############

%################3：自己的算法为什么没有这些问题，哪些结构或者设计规避掉了这些问题(详细写)#########
% These findings underscore the superior performance of our proposed structure, SRS, in CT tasks. SRS not only achieves the highest accuracy but also showcases improvements in each kind of viscus such as Aorta, Gallbladde. Considering its advantages, such as the ability to perform segmentation effectively across different reflective media and achieving state-of-the-art performance, these two factors further highlight the applicability of our network, ensuring effective medical image segmentation.
%################3：自己的算法为什么没有这些问题，哪些结构或者设计规避掉了这些问题(详细写)#########

\subsection{Experiments on Infrared Images}

\label{Section:EII}

%################1：目前现象是什么(详细写，结合数据定量说明一下好在哪里)。2：相对于自己的算法，能看出来别人的算法有什么问题(详细写，或者说出现的原因)，3：自己的算法为什么没有这些问题，哪些结构或者设计规避掉了这些问题(详细写).################
% \begin{figure*}[!t]
%   \centering
%    \includegraphics[width=0.7\linewidth]
%    {NetworkStructure/label_image/supply_other2.pdf}
%    \caption{Visual comparison of results produced by DNANet~\cite{DNANET}, ALCNet~\cite{ALCNET} and APGCNet~\cite{AGPCNET} on infrared image dataset. Row 1, 2 NUDT - KBT2019 samples, Row 3, 4 and 5 - SIRST samples. Yellow boxes represent error segmentation.}
%       \label{fig:supply_infrar}
% \end{figure*}
%################1：目前现象是什么(详细写，结合数据定量说明一下好在哪几个方面)###############
In contrast to medical image datasets, infrared image datasets exhibit more severe challenges, including internal feature scarcity, minimal target pixel occupancy (potentially just a few pixels), and complex backgrounds. Therefore, the segmentation of weak infrared small targets presents a greater challenge to network performance. 

% In addition to the four state-of-the-art neural network algorithms mentioned earlier, Tab.~\ref{tab:ExOnISTDataset} displays IoU and F1 scores for these algorithms. Also, Fig.~\ref{fig:supply_infrar} shows that SRS achieves the highest segmentation performance on both the SIRST and NUDT-KBT19 datasets. Compared with the current state-of-the-art algorithm DNANet, SRS demonstrates a 1.54\% improvement on SIRST and a 0.68\% improvement on NUDT-KBT19. This indicates that SRS can achieve state-of-the-art results in the field of infrared small target segmentation as well.
Tab.~\ref{tab:ExOnISTDataset} shows the the quantitative evaluation of the results produced by both the natural image segmentation network and the networks specifically designed for infrared images. The experiments show that SRS achieves the best segmentation performance on both the SIRST and NUDT-KBT19 datasets. Compared with the current state-of-the-art algorithm DNANet, SRS shows a 1.54\% improvement in SIRST and a 0.68\% improvement in NUDT-KBT19. Regardless of the dataset taken in various reflective mediums, SRS can effectively make a segmentation of the target area and achieves the state-of-the-art performance.

\subsection{Zero-shot Segmentation}
{We conduct domain generalization experiments. Specifically, models pre-trained on a source domain within one modality are evaluated under a zero-shot setting on unseen modalities. We select the four modalities mentioned: Breast Ultrasound (BUSI → BUS), Thyroid Ultrasound (TNSCUI → TUCC), Skin Cancer (ISIC → PH2), and Infrared Small Target (SIRST → IRSTD-1K). As shown in Table~\ref{tab:zeroshot}, SRSNet achieves IoU improvements of 1.4\%, 2.14\%, and 7.39\% over the second-best method on the Breast Ultrasound, Skin Cancer, and Infrared Small Target modalities, respectively. This demonstrates the strong generalization capability of our method on unseen datasets.}

{Furthermore, as presented in Table~\ref{tab:ExOnUlt_Der} and Table.~\ref{tab:ExOnISTDataset}, networks with relatively lower parameters and GFLOPs are typically specialized for single modalities, such as EGE-Unet for skin cancer or SwinUnet for CT. These models generally exhibit significantly degraded performance on other modalities; for instance, EGE-Unet performs well below its typical level on Breast Ultrasound (BUS) and CT. In contrast, networks demonstrating robust performance across diverse medical and infrared modalities—such as CMU-Net\cite{CMUNET}, EMCAD\cite{rahman2024emcad}, and DNANet\cite{DNANET}—tend to have higher parameter counts and computational complexity. However, this design affords greater generality, eliminating the need for modality-specific network design. Crucially, our network currently achieves state-of-the-art performance across all evaluated modalities, performing comparably to or even surpassing networks specifically designed for individual modalities. This further enhances the practical value of our SRS approach.}

\subsection{Bottleneck of SRS}

{As shown in Fig.~\ref{fig:failure_case}, we present failure cases for breast cancer (BUS, BUSI) (first and second columns), malignant thyroid cancer (TNSCUI) (third column), and infrared small target (SIRST) (fourth and fifth columns), respectively. We employ Grad-CAM\cite{selvaraju2017grad} to visualize the last layer of the encoder, which is the layer where dynamic convolution is located.
Taking the medical image of breast cancer as an example, in breast ultrasound images, the acoustic impedance difference between normal tissue and tumor tissue is minimal, and the density of tumors is extremely close to that of normal tissue. This results in indistinct pixel value differences within the tumor and weak gradients. Although our DPConv (as shown in Fig. d) brings attention focus to tumor edges (where significant gradient differences exist with surrounding tissues) and reduces attention misclassification, thereby encouraging the network to focus more on the target region, the SRS network itself adopts dense nested and skip-connection structures, which introduce additional attention (yellow box regions), consequently leading to segmentation failure. This phenomenon is also evident in the third column for malignant thyroid cancer.}

{For infrared small targets, the network segmentation performance also relies on gradient differences for infrared weak small target segmentation. However, targets often appear as noise-like forms. The segmentation failure in the fourth column is similar to that in medical images, where the attention of DPConv itself is relatively accurate and concentrated near the target, but the SRS network itself introduces interfering attention.
The segmentation failure in the fifth column occurs because the target itself is excessively small and appears as noise. Although gradients exist, DPConv cannot separate it from the noise in the yellow box region. This is because reconstruction employs MSE loss, which focuses more on the overall structure rather than individual points. Specifically, the features transmitted from phase 1 do not focus on infrared small targets that appear as noise points. Consequently, the constructed DPConv cannot effectively attend to infrared small target regions.}

\begin{table}[!t]
\centering
\caption{Ablation Study on Reconstruction Methods. R and J mean respective and joint. Best results are highlighted as \cellcolor{l}{\bf \!first\!}, \cellcolor{ll}{\!second\!} .}
\label{tab:reconstructionmethod}
\renewcommand{\arraystretch}{1.0}
\setlength{\tabcolsep}{3pt}
\scriptsize
\begin{tabular}{@{}lcc cc cc cc cc }
\toprule
\multirow{4}{*}{{Method}} & \multirow{4}{*}{{Str}} & \multirow{4}{*}{{Data}} & \multicolumn{8}{c}{{Performance}} \\
\cmidrule(l){4-11}
&& & \multicolumn{2}{c}{{BUS}} & \multicolumn{2}{c}{{BUSI}} & \multicolumn{2}{c}{{SIRST}} & \multicolumn{2}{c}{{Avg}} \\
\cmidrule(lr){4-5} \cmidrule(lr){6-7} \cmidrule(lr){8-9} \cmidrule(l){10-11}
&& & {IoU} & {F1} & {IoU} & {F1} & {IoU} & {F1} & {IoU} & {F1} \\
\midrule
\multirow{4}{*}{{Spark}} & w/o F & R & 88.55 & 93.69 & 75.83 & 84.05 & 75.46 & 85.95 & 79.94 & 87.89 \\
& Frz & R & 88.59 & 93.75 & \cellcolor{ll}{75.89} & \cellcolor{ll}{84.25} & 75.52 & 85.99 & 80.00 & \cellcolor{ll}{87.99} \\
& w/o F & J & \cellcolor{ll}{88.81} & \cellcolor{ll}{93.86} & 75.74 & 83.86 & \cellcolor{ll}{75.60} & 85.98 & \cellcolor{ll}{80.05} & 87.90 \\
& Frz & J & \cellcolor{l}{\textbf{88.96}} & \cellcolor{l}{\textbf{94.00}} & \cellcolor{l}{\textbf{76.01}} & \cellcolor{l}{\textbf{84.37}} & \cellcolor{l}{\textbf{75.63}} & \cellcolor{l}{\textbf{86.01}} & \cellcolor{l}{\textbf{80.20}} & \cellcolor{l}{\textbf{88.12}} \\
\hline
\multirow{4}{*}{Direct} & w/o F & R & 88.23 & 92.87 & 74.60 & 82.93 & 74.61 & 84.19 & 79.14 & 86.66 \\
& Frz & R & 88.29 & 93.45 & 74.47 & 82.57 & 74.67 & 84.28 & 79.14 & 86.76 \\
& w/o F & J & 88.29 & 93.01 & 74.62 & 83.01 & 74.65 & 84.30 & 79.18 & 86.77 \\
& Frz & J & 88.27 & 93.44 & 74.52 & 82.64 & 74.71 & 84.33 & 79.16 & 86.80 \\
\bottomrule
\end{tabular}
\end{table}

\section{Ablation Study}
\subsection{Reconstruction Method}
{We select two reconstruction method: Direct reconstruction and Spark pretraining reconstruction\cite{spark}. Direct reconstruction refers to the method where the input image is directly fed into the encoder and then reconstructed to generate the reconstructed image. And Spark is a sparse mask modeling approach for convolutional neural networks applicable to pre-training any CNN architecture to learn robust local representations. This method is applied to our reconstruction network. During the reconstruction phase, Spark first masks image patches, requiring the network to reconstruct masked regions from visible areas, thereby enabling the reconstruction encoder to learn more robust features.
As demonstrated in Table~\ref{tab:reconstructionmethod}, the reconstruction network utilizing Spark pretraining yields significantly enhanced benefits for DPConv construction. }

{We also investigate two critical factors affecting DPConv performance: Spark pretraining dataset scale and encoder freezing strategies.}

\begin{enumerate}
    \item \textbf{Dataset Scale Analysis}: {We expand the pretraining dataset using two configurations. "Respectively" denotes dataset-specific pretraining, where encoders pretrained on individual datasets (e.g., BUS) are exclusively applied to corresponding segmentation tasks. "Combine" represents joint pretraining, where medical images utilize BUS, BUSI, and TNSCUI datasets, while infrared images employ SIRST, HIT-UAV, LLVIP,
    NUDT-KBT datasets for pretraining across all segmentation tasks. As Avg IoU shown in Table.~\ref{tab:reconstructionmethod} (80.20 vs 88.00), expanding the dataset can significantly improve the performance, which further highlights the potential of our approach.}
    \item \textbf{Encoder Freezing Impact}: {We examine the influence of encoder freezing on downstream segmentation tasks. Following Spark pretraining, encoder freezing produces significant effects. Fine-tuning the encoder facilitates superior DPConv generation and enhanced segmentation performance, which can be demonstrated by Avg IoU in Table.~\ref{tab:reconstructionmethod} (80.20 vs 80.05)}
\end{enumerate}

\begin{figure}
    \centering
    \includegraphics[width=0.7\linewidth]{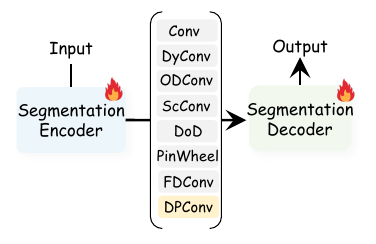}
    \caption{The illustration of the ablation of DPConv. The segmentation network used in phase two is selected as the baseline.}
    \label{fig:ablationofdpconv}
\end{figure}

\begin{table*}[h]
\centering
\caption{Quantitative evaluation of the results. Best results are highlighted as \colorbox{l}{\bf \!first\!}, \colorbox{ll}{\!second\!}.\label{tab:ExOnDyConv}}
\resizebox{\textwidth}{!}{%
\begin{tabular}{lccccccccccccc}
\hline
\multirow{4}{*}{Convolution} & \multirow{4}{*}{{Publication}}  & \multicolumn{2}{c}{{Computation}}&\multicolumn{10}{c}{{Performance}} \\

\cmidrule(lr){3-4} \cmidrule(l){5-14}
& & \multirow{3}{*}{{Params (M)}} & \multirow{3}{*}{{GFLOPs (G)}} & \multicolumn{2}{c|}{{BUS}} & \multicolumn{2}{c|}{{BUSI}} & \multicolumn{2}{c|}{{TNSCUI}} & \multicolumn{2}{c|}{{SIRST}} & \multicolumn{2}{c}{{Avg}} \\

\cmidrule(lr){5-6} \cmidrule(lr){7-8} \cmidrule(lr){9-10} \cmidrule(lr){11-12} \cmidrule(lr){13-14}
& & & & {IoU} & {F1} & {IoU} & {F1} & {IoU} & {F1} & {IoU} & {F1} & {IoU} & {F1} \\

\midrule
Traditional Conv & - & 9.17 & 47.461 & {87.42} & {93.53} & {73.85} & {82.00} & {78.34} & {86.13} & {72.34} & {83.03} & {77.98} & {86.17} \\[2pt]
DyConv \cite{DYCONV} & CVPR'20 & 9.19 & 47.461 & {87.60} & {93.16} & {71.35} & {79.80} & {78.12} & {86.01} & {71.13} & {81.70} & {77.05} & {85.16} \\[2pt]
ODConv \cite{ODCONV} & ICLR'22 & 9.17 & 47.461 & {87.58} & {93.16} & {70.82} & {79.37} & {78.26} & {86.15} & {71.70} & {82.97} & {77.09} & {85.41} \\[2pt]
ScConv \cite{ScConv} & ICLR'23 & 9.27 & 47.586 & {87.43} & {93.02} & {71.29} & {79.85} & {78.59} & {86.33} & {71.03} & {81.83} & {77.08} & {85.25} \\[2pt]
DoD\cite{xie2023learning} & TPAMI'23 & 8.68 & 45.178 & {86.85} & {92.62} & {72.41} & {81.38} & {78.41} & {86.25} & {68.58} & {80.12} & {76.56} & {85.09} \\[2pt]
{PinWheel}\cite{yang2025pinwheel} & AAAI'25 & 9.05 & 46.42 & {87.78} & \cellcolor{ll}{93.49} & {74.05} & {81.95} & {79.01} & {86.69} & {71.80} & {82.59} & {78.16} & {86.18} \\[2pt]
{FDConv}\cite{chen2025frequency} & CVPR'25 & 8.72 & 45.94 & {87.97} & {93.38} & {73.92} & \cellcolor{ll}{82.80} & {79.04} & {86.74} & {72.75} & {83.66} & {78.42} & {86.64} \\
\midrule
DPConv(SRS) & & 10.35 & 47.467 & \cellcolor{ll}{88.29} & {93.44} & \cellcolor{ll}{74.47} & {82.57} & \cellcolor{ll}{79.33} & \cellcolor{ll}{87.02} & \cellcolor{ll}{74.67} & \cellcolor{ll}{84.28} & \cellcolor{ll}{79.19} & \cellcolor{ll}{86.82} \\
{DPConv-Spark} & & 10.35 & 47.467 & \cellcolor{l}{\textbf{88.81}} & \cellcolor{l}{\textbf{93.86}} & \cellcolor{l}{\textbf{75.74}} & \cellcolor{l}{\textbf{83.86}} & \cellcolor{l}{\textbf{80.25}} & \cellcolor{l}{\textbf{88.96}} & \cellcolor{l}{\textbf{75.60}} & \cellcolor{l}{\textbf{85.98}} & \cellcolor{l}{\textbf{80.10}} & \cellcolor{l}{\textbf{88.17}} \\
\hline
\end{tabular}%
}
\end{table*}

\subsection{DPConv}

In Tab.~\ref{tab:ExOnDyConv}, we conduct ablation experiments on DPConv, considering ScConv\cite{ScConv}, DyConv\cite{DYCONV}, ODConv\cite{ODCONV} and DOD\cite{xie2023learning} as alternative choices. The segmentation network used in phase two is selected as the baseline. Fig.~\ref{fig:ablationofdpconv} illustrates the network configuration of the ablation study mentioned above. We replace DPConv with other dynamic convolution, and the other parts of the segmentation network are completely identical.

As shown in Tab.~\ref{tab:ExOnDyConv}, it is evident that the selection of dynamic convolution has a substantial impact on the overall performance of the model. The results demonstrate that DPConv, as introduced in this paper, outperforms traditional dynamic convolutions by a large margin. For example, when tested on the BUSI dataset, DPConv shows an improvement of 2.88\% in IoU and 2. 72\% in the F1 score compared to ScConv\cite{ScConv}. 

Moreover, as discussed earlier, dynamic convolutions designed for natural images encounter significant challenges when applied to weak target images. This is validated through the results in Tab.~\ref{tab:ExOnDyConv}, which shows that the segmentation networks utilizing dynamic convolutions exhibit worse segmentation performance compared to traditional conv. Compared with traditional conv, DyConv, ODConv, ScConv and DOD show a decline of 0.94\%, 0.89\%, 0.90\% and 1.42\% in Avg IoU, respectively. However, DPConv overcomes this limitation and improves the precision of segmentation by 1.21\%.
 
\subsection{Reconstruction Network of SRS}

\begin{figure*}[!t]
\centering
\includegraphics[width=1.0\linewidth]{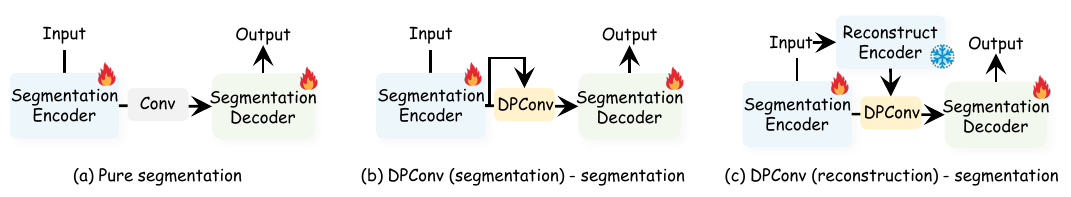}
\caption{The illustration of the ablation of the reconstruction task. The experiments (a)-(c) and (b)-(c) respectively validate the effect of reconstruction in the SRS and DPConv.}
\label{fig:albationofreconstruction}
\end{figure*}

As shown in Fig.~\ref{fig:albationofreconstruction}, we design two experiments to validate the effect of the reconstruction network in SRS and DPConv. we define three network architectures: Pure segmentation in Fig.~\ref{fig:albationofreconstruction}(a), DPConv (segmentation)-segmentation in Fig.~\ref{fig:albationofreconstruction}(b), DPConv (reconstruction)-segmentation in Fig.~\ref{fig:albationofreconstruction}(c).
Pure segmentation utilizes the segmentation network mentioned in phase two. DPConv (reconstruction)-segmentation means that the DPConv kernel is generated from the segmentation network. DPConv (reconstruction)-segmentation means that the DPConv kernel is generated from the reconstruction network. We design the first experiment as the comparison of Fig.~\ref{fig:albationofreconstruction}(a) with Fig.~\ref{fig:albationofreconstruction}(c) to show the effect of reconstruction in the SRS.
The second experiment is designed as the comparison of Fig.~\ref{fig:albationofreconstruction}(b) with Fig.~\ref{fig:albationofreconstruction}(c) to show the effect of the condensed semantic feature offered by the reconstruction network for the generation of DPConv kernel.

Tab.~\ref{tab:ablation1} shows the experimental results of the above configurations.
In the first experiment, the result shows that the reconstruction improves the IoU by 1.27\% on the average. This indicates that the reconstruction task can effectively utilize DPConv to transfer beneficial domain knowledge to the segmentation task, thus improving the representational capacity of the network. In the second experiment, the DPConv kernel generated from the reconstruction network has an improvement of 1.02\% on IoU on average. This shows that the reconstruction task is also important in the generation of DPConv kernel.

\begin{table*}[!t]
\centering
\caption{Ablation Study on Reconstruction Network Components. Best results are highlighted as \colorbox{l}{\bf \!first\!}, \colorbox{ll}{\!second\!}.}
\label{tab:ablation1}
\renewcommand{\arraystretch}{1.4}
\setlength{\tabcolsep}{6pt}
\begin{tabular}{@{}l@{\hspace{8pt}}ccc@{\hspace{12pt}}cc@{\hspace{8pt}}cc@{\hspace{8pt}}cc@{\hspace{8pt}}cc@{}}
\toprule
\multirow{3}{*}{{Network Configuration}} & \multicolumn{3}{c@{\hspace{12pt}}}{{Component Analysis}} & \multicolumn{8}{c}{{Performance Metrics}} \\
\cmidrule(lr){2-4} \cmidrule(l){5-12}
& {Seg.} & {Kernel} & {Kernel} & \multicolumn{2}{c@{\hspace{8pt}}}{{BUS}} & \multicolumn{2}{c@{\hspace{8pt}}}{{BUSI}} & \multicolumn{2}{c@{\hspace{8pt}}}{{SIRST}} & \multicolumn{2}{c}{{Average}} \\
& {Net} & {(Recon.)} & {(Seg.)} & {IoU} & {F1} & {IoU} & {F1} & {IoU} & {F1} & {IoU} & {F1} \\
\midrule
Pure Segmentation & \checkmark & \ding{55} & \ding{55} & \cellcolor{ll}{87.42} & \cellcolor{ll}{93.53} & \cellcolor{ll}{73.85} & {82.00} & {72.34} & {83.03} & {77.87} & {86.18} \\
\addlinespace[2pt]
DPConv (Seg.) + Segmentation & \checkmark & \ding{55} & \checkmark & {86.98} & {92.68} & {73.72} & \cellcolor{ll}{82.04} & \cellcolor{ll}{73.67} & \cellcolor{ll}{84.28} & \cellcolor{ll}{78.12} & \cellcolor{ll}{86.33} \\
\addlinespace[2pt]
DPConv (Recon.) + Segmentation & \checkmark & \checkmark & \ding{55} & \cellcolor{l}{\textbf{88.29}} & \cellcolor{l}{\textbf{93.45}} & \cellcolor{l}{\textbf{74.47}} & \cellcolor{l}{\textbf{82.57}} & \cellcolor{l}{\textbf{74.67}} & \cellcolor{l}{\textbf{84.28}} & \cellcolor{l}{\textbf{79.14}} & \cellcolor{l}{\textbf{86.76}} \\
\bottomrule
\end{tabular}
\end{table*}

%% file: Section/Conclusion.tex
\section{Limitation and Future work}
{Although the proposed SRSNet achieves state-of-the-art performance in weak target segmentation domains (medical and infrared), our network still encounters several limitations. Since DPConv generation requires manipulation of input feature channels, the parameter count of the DPConv generation subnetwork expands dramatically when the number of channels becomes excessive. Additionally, due to the extensive use of AdaptivePooling, tensor concatenation, and other operations throughout the network, it suffers from relatively low real-time computational efficiency and cannot achieve ideal inference speed. Furthermore, we discover that DPConv performance is strongly correlated with the upstream reconstruction task methodology. Adopting the Spark reconstruction method brings greater improvements compared to direct reconstruction. Moreover, dataset expansion also enhances DPConv performance. Currently, we have only expanded the dataset by 2-3 times, which falls far short of the level achieved in Spark (100 times). We will explore the aforementioned issues in future work and further optimize DPConv.}
\section{Conclusion}
\label{Sec:Conclusion}

In this paper, we analyze the challenge of existing dynamic convolutions in medical image segmentation and infrared image segmentation and propose a new type of dynamic convolution called dynamic parameter convolution (DPConv). Specifically, DPConv can efficiently leverage deep features from the reconstruction network to generate DPConv kernel which can enhance the performance of the segmentation network. We denote the the segmentation network integrated with DPConv generated from reconstruction network as the siamese reconstruction-segmentation network. We show that our DPConv can properly bridge the reconstruction task and the segmentation task, and the performance of segmentation can be greatly boosted by simply training our DPConv on the reconstruction task with unlabeled data. Comprehensive experimental results demonstrate the superior performance of our method over the recent state-of-the-art approaches.

\section{Acknowledgments}
Supported by Natural Science Foundation of China under
Grant 62376153, 62402318, 24Z990200676, 62272419, 62271465. Suzhou
Basic Research Program under Grant SYG202338. Joint Funds of the National Natural Science Foundation of China (U22A2033)

% \section{References}
% \bibliographystyle{unsrt}
% \bibliography{main}

% \newpage

%% file: Section/authors.tex
 
\vskip-40pt % 也可以使用 \vskip-40pt 等具体数值
\begin{IEEEbiography}[{\includegraphics[width=0.75in,height=1in,clip]{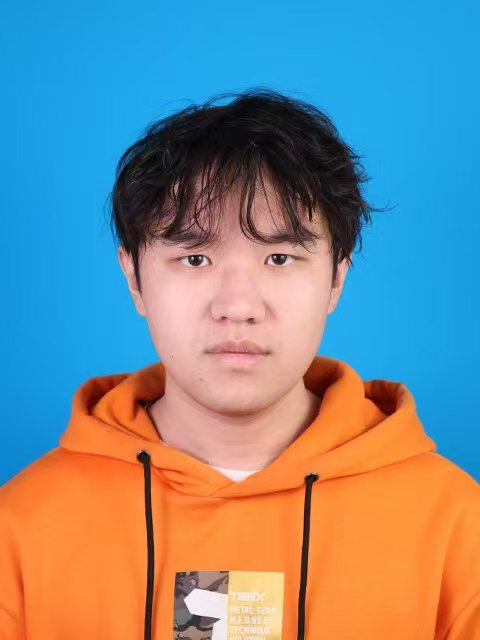}}]{Bingkun Nian} 
received B.S. degree from Harbin Institute of Technology, China, 2021. Now, He is pursuing Ph.D in Shanghai Jiao Tong University.
His current interests include weak target detection, deep learning, medical image segmentation. 
\end{IEEEbiography}
\vskip-40pt % 也可以使用 \vskip-40pt 等具体数值

\begin{IEEEbiography}[{\includegraphics[width=0.75in,height=1in,clip]{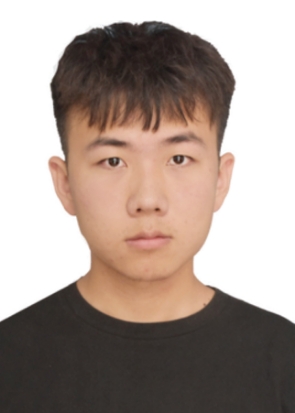}}]{Fenghe Tang} received the B.Eng. and M.Eng. degrees from Harbin Institute of Technology (HIT) in 2021 and 2023, respectively. He is currently pursuing the Ph.D. degree at the University of Science and Technology of China (USTC). His research interests include foundation models and medical image analysis. He has published papers in in top-tier journals and conferences, such as Nature Communications, npj Digital Medicine, Medical Image Analysis, CVPR, ACM MM, and MICCAI.
\end{IEEEbiography}
\vskip-40pt % 也可以使用 \vskip-40pt 等具体数值

\begin{IEEEbiography}[{\includegraphics[width=0.75in,height=1in,clip]{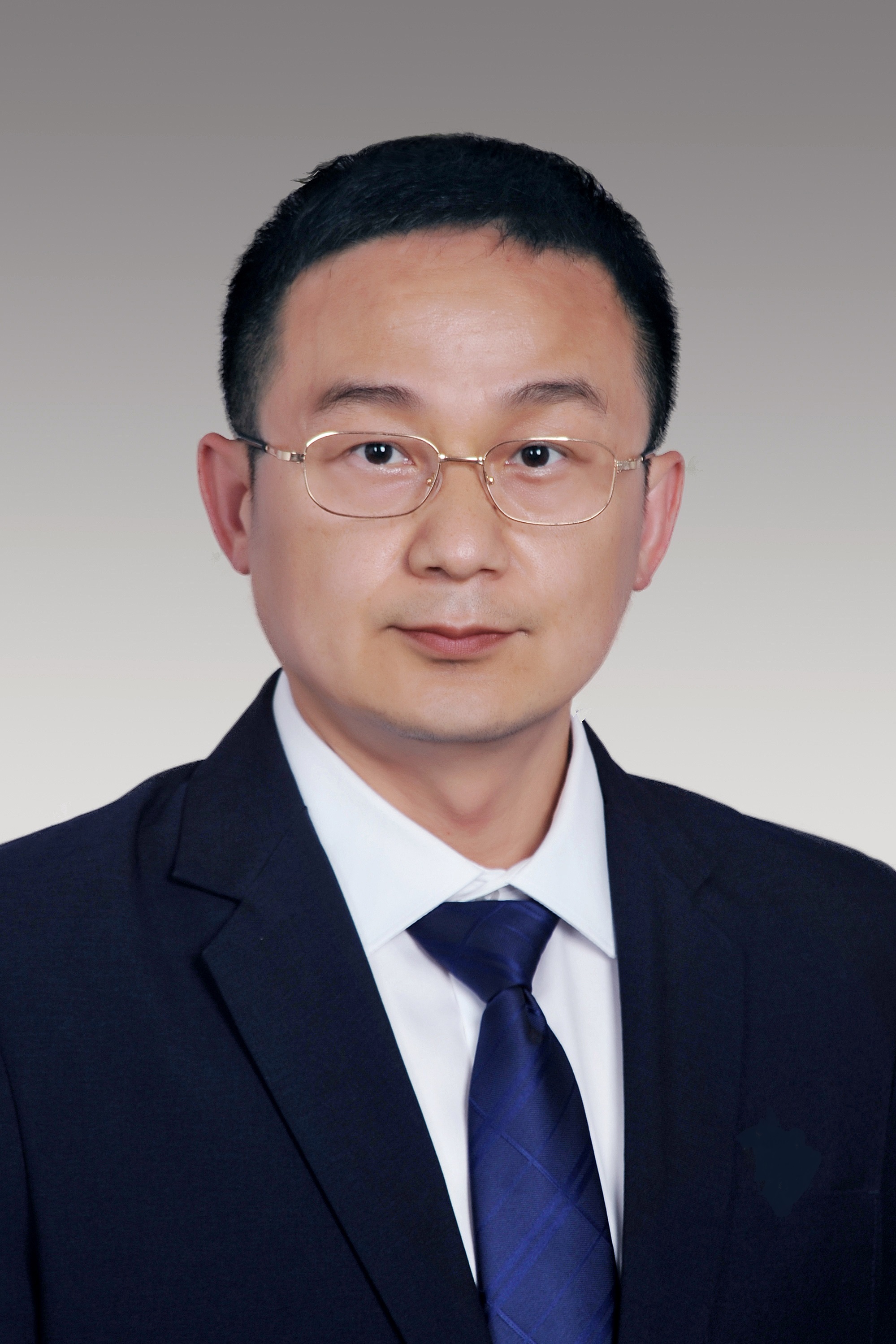}}]{Jianrui Ding} received the B.Eng. degree and the Ph.D. degree from the University of Harbin Institute of Technology, China, in 1995 and 2013. He is currently an Associate Professor and the Head of the Department of Artificial Intelligence, School of Computer Science and Technology, Harbin Institute of Technology (Weihai). His research interests include artificial intelligence, computer vision, medical image analysis, and their applications.
\end{IEEEbiography}

\vskip-40pt % 也可以使用 \vskip-40pt 等具体数值
\begin{IEEEbiography}[{\includegraphics[width=0.75in,height=1in,clip]{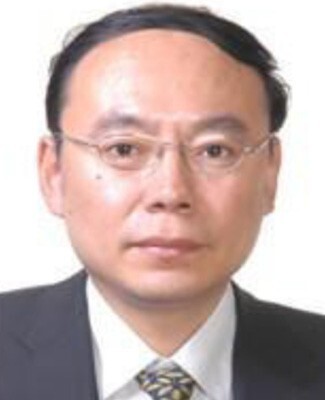}}]{Jie Yang} received his PhD from the Department of Computer Science, Hamburg University, Germany, in 1994. Currently, he is a professor at the Institute of Image Processing and Pattern Recognition, Shanghai Jiao Tong University, China. He has led many research projects (e.g.,National Science Foundation, 863 National High Tech. Plan), had one book published in Germany, and authored more than 200 journal papers. His major research interests are object detection and recognition, data fusion and data mining, and medical image processing.
\end{IEEEbiography}
\vskip-40pt % 也可以使用 \vskip-40pt 等具体数值
\begin{IEEEbiography}[{\includegraphics[width=0.75in,height=1in,clip]{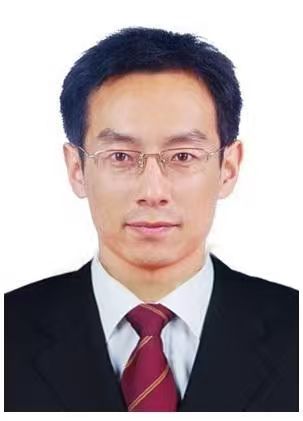}}]{Zhonglong Zheng} received the B.Eng. degree from
the University of Petroleum, China, in 1999, and the Ph.D. degree from Shanghai Jiaotong University, China, in 2005. He is currently a Full Professor and the Dean of the School of Computer Science, Zhejiang Normal University, China. His research interests include machine learning, computer vision, and the IoT technology with their applications.
\end{IEEEbiography}
\vskip-40pt % 也可以使用 \vskip-40pt 等具体数值
\begin{IEEEbiography}
[{\includegraphics[width=0.75in,height=1in,clip]{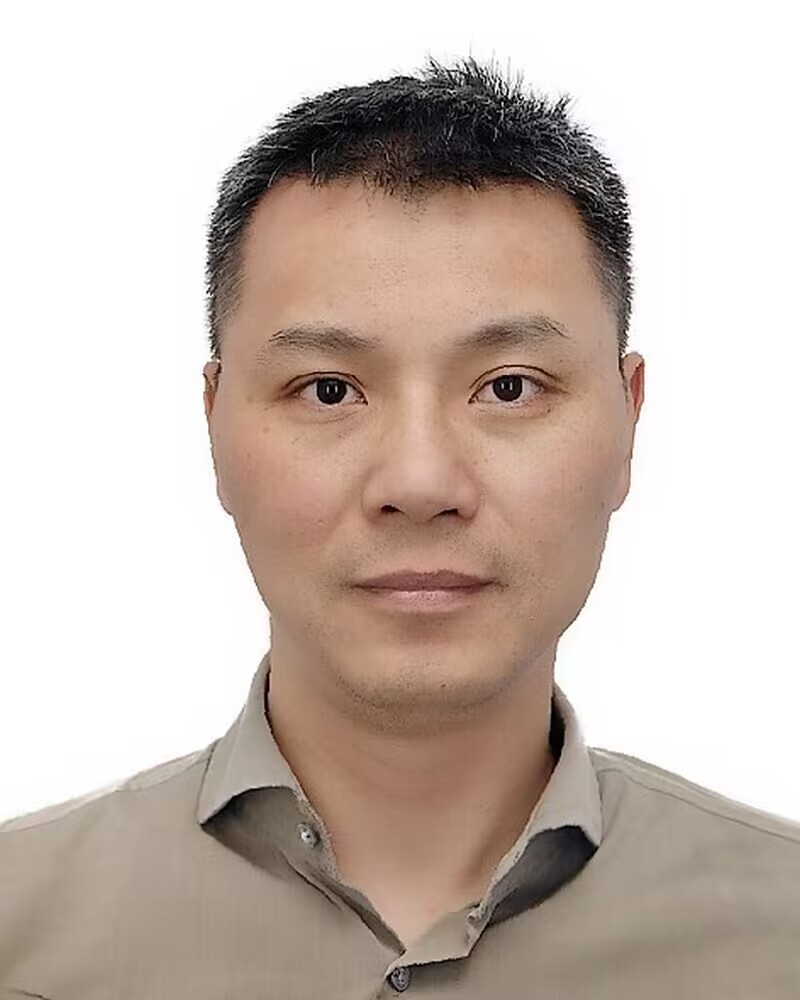}}]{Shaohua Kevin Zhou} is dedicated to research on medical
image computing, especially analysis and reconstruction,
and its applications in real practices.
He obtained his Bachelor’s degree from University
of Science and Technology of China (USTC),
Master’s degree from National University of Singapore
(NUS), and PhD degree from University of
Maryland (UMD), College Park. Currently he is a
distinguished professor and executive dean of School
of Biomedical Engineering, Suzhou Institute for
Advanced Research, USTC and directs the Center
for Medical Imaging, Robotics, Analytic Computing \& Learning (MIRACLE).
He was a principal expert and a senior R\&D director at Siemens
Healthcare Research. He is an editorial board member for IEEE Transactions
on Medical Imaging (16-23), IEEE Transactions on Pattern Analysis and Machine
Intelligence, Medical Image Analysis, and npj Digitial Medicine, and an area chair for AAAI, CVPR, ICCV, MICCAI, and NeurIPS. He is a Fellow of AIMBE, IAMBE, IEEE, MICCAI, and NAI.
\end{IEEEbiography}
\vskip-40pt % 也可以使用 \vskip-40pt 等具体数值
\begin{IEEEbiography}[{\includegraphics[width=0.75in,height=1in,clip]{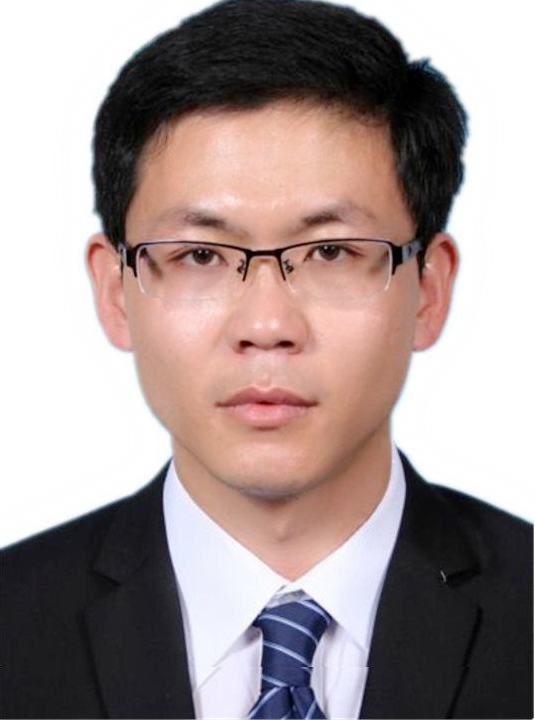}}]{Wei Liu} received the B.S. degree from Xi'an Jiaotong University, Xi'an, China, in 2012. He received the Ph.D. degree from Shanghai Jiao Tong University, Shanghai, China in 2019. He was a research fellow in The University of Adelaide and The University of Hong Kong from 2018 to 2021 and 2021 to 2022, respectively. He has been working as an associate professor in Shanghai Jiao Tong University since 2022. His current research areas include medical image anaylysis, 3D detection, 3D reconstruction and self-supervised depth estimation.
\end{IEEEbiography}